# Perception and Sensing for Autonomous Vehicles Under Adverse Weather Conditions: A Survey

Yuxiao Zhang[a,*], Alexander Carballo[b,c], Hanting Yang[a] and Kazuya Takeda[a,b,c]

[a]*Graduate School of Informatics, Nagoya University, Nagoya, 464-8601, Japan*
[b]*Institute of Innovation for Future Society, Nagoya University, Nagoya, 464-8601, Japan*
[c]*TierIV Inc., Nagoya University Open Innovation Center, 1-3, Mei-eki 1-chome, Nakamura-Ward, Nagoya, 450-6610, Japan*

ABSTRACT

Automated Driving Systems (ADS) open up a new domain for the automotive industry and offer new possibilities for future transportation with higher efficiency and comfortable experiences. However, autonomous driving under adverse weather conditions has been the problem that keeps autonomous vehicles (AVs) from going to level 4 or higher autonomy for a long time. This paper assesses the influences and challenges that weather brings to ADS sensors in an analytic and statistical way, and surveys the solutions against inclement weather conditions. State-of-the-art techniques on perception enhancement with regard to each kind of weather are thoroughly reported. External auxiliary solutions, weather conditions coverage in currently available datasets, simulators, and experimental facilities with weather chambers are distinctly sorted out. Additionally, potential future ADS sensors candidates and approaches beyond common senses are provided. By looking into all kinds of major weather problems the autonomous driving field is currently facing, and reviewing both hardware and computer science solutions in recent years, this survey points out the main moving trends of adverse weather problems in autonomous driving, i.e., advanced sensor fusions, more sophisticated networks, and V2X & IoT technologies; and also the limitations brought by emerging 1550 nm LiDARs. In general, this work contributes a holistic overview of the obstacles and directions of ADS development in terms of adverse weather driving conditions.

## 1. Introduction

Autonomous Vehicles (AVs) and Automated Driving Systems (ADS) are the frontiers of today's automotive. Ever since the birth of automobiles, no technology has changed cars in such a revolutionary way. Autonomous vehicles bring fewer traffic accidents and fatalities, lessened energy consumption and air pollution, and increased access to transportation for people with limited reliable mobility options. Driverless vehicles are changing the way people and goods are transported fundamentally and could benefit the future society in significant ways. However, incidents and casualties involving vehicles equipped with ADS are still disturbingly rising. For the merits of autonomous vehicles to be recognized more extensively, the immediate problem of ADS must be appropriately dealt with, namely, the performance of autonomous cars in adverse weather conditions [1] [2] [3].

Weather phenomena have various negative influences on traffic and transportation. Averagely, global precipitation occurs 11.0% of the time [4]. It has been proven that the risk of accidents in rain conditions is 70% higher than normal [5]. 77% of the countries in the world receive snow. Take the United States national statistics as an example, each year over 30,000 vehicle crashes occur on snowy or icy roads or during snowfall or sleet [6], so the threat from snow is bona fide. Phenomena like fog, haze, sandstorms, and strong light severely decrease the visibility and the difficulties they cause for driving are palpable [7]. Secondary problems directly or circumstantially caused by weather, like heat & coldness, contamination, or damage of vehicle hardware, also have unpredictable or undesirable effects on both manned and autonomous cars.

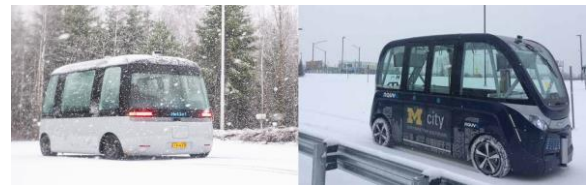

(a) Sensible4       (b) Mcity

**Figure 1:** (a) Sensible4 autonomous bus in snow. Image courtesy of Mr. Tsuneki Kaiho, Sensible4. (b) Mcity level 4 self-driving shuttle. Image courtesy of Dr. Huei Peng, University of Michigan.

With some rapid development during recent years, there are already many deployable autonomous cars either on trial stage or in operation all over the world, and with the help of LiDAR (Light Detection And Ranging, sometimes Light Imaging Detection And Ranging for the image-like resolution of modern 3D sensors) technology, some manufacturers claim to have achieved or about to deliver vehicles with autonomy equivalent to level 4 of SAE standard [8] like Waymo's commercial self-driving taxi service in Phoenix, Arizona [9]. In the past few years, the University of Michigan successfully held a driver-less shuttle project called Mcity on the Campus of Ann Arbor as the first Level 4 automated shuttle project in the United States as shown in Fig. 1(b) [10]. However, an inevitable problem for all the current autonomous cars is that they barely operate during heavy rain or snow due to safety issues. Even though lots of research and tests have been conducted in adverse weather conditions, the Mcity shuttle would be shut down for operation when the shuttle's windshield wipers need to run continuously in rain or snow. The Sohjoa Baltic Project [11] reveals that the autonomous mini bus failed to charge properly overnight in the Estonian winter because of the low temperature, and the daily operating hours had to be decreased due to the extra power consumption for heating. On the other hand, Sensible4 from Finland does not stop on snow and has already begun open test drives in snowy weather as shown in Fig.1(a) [12].

---

*Corresponding author
yuxiao.zhang@g.sp.m.is.nagoya-u.ac.jp (Y. Zhang)
ORCID(s): 0000-0002-8082-301X (Y. Zhang)

Looking at what we already have in hand right now, it's worth noticing that most of the major automotive enterprises are planning on skipping level 3 autonomy (conditional automation) and leaping to level 4 directly, due to the potential accident liability issue in law and the necessity of car-human handover, leaving Audi A8 and Honda the minority players who are declaring commitment in level 3 right now in the general market [13]. What's clear for sure is that level 2 autonomy, or to say multiple Advanced Driver Assistance Systems (ADAS), mostly manifested on adaptive cruise control and hands-on/free lane-centering feature, can be offered by almost every mainstream manufacturer in the auto industry [14]. Nevertheless, the performance of level 2 in rainy or snowy conditions barely meets the expectation, as shown in Fig.2 where the lane-keeping feature oversteered a car during skidding in snow on a highway. Tesla's autopilot can somehow manage to navigate through normal rain or snow with road marks clear in sight but still struggles in certain tricky situations, like a heavy storm or covered lane lines [15]. As for the other typical level 2 provider, General Motor's Super Cruise, the use of the self-driving function is officially prohibited in slippery or in other adverse conditions, including rain, sleet, fog, ice, or snow [16]. Apparently, adverse weather conditions are restraining human drivers to the steering wheel and AVs still can not be fully trusted to be alone yet. As a result, for ADS to continue moving forward to the next era, autonomous cars need more to get past all the weather.

There have been lots of researchers from all over the world working on a particular sensor's better ability in dealing with rain and fog, few working on snow. Besides overviews on the driveability assessment in autonomous driving [17], there are literature reviews on common sensors' performance evaluations used in ADS in weather conditions [18] [19] [20]. There is no paper right now that has covered all the weather phenomena, all the direct and secondary impacts on AVs, different AV components' performance in adverse weather conditions, sensor solutions on both hardware and software sides, perception enhancement methods for each weather kind, and potential solutions with other technologies, in a comprehensive way. So, in addition to

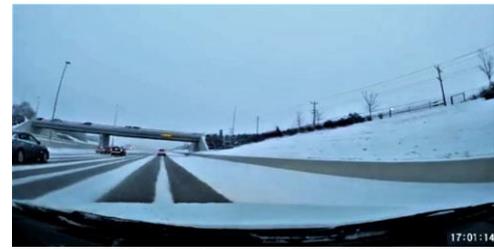

(a) Driving on highway in snow when skidding happened.

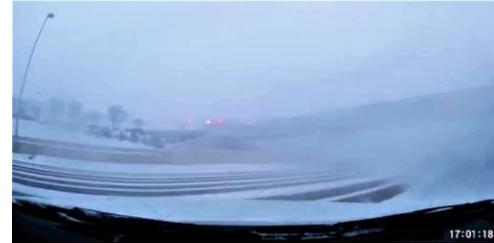

(b) Lane keeping feature oversteered car on slippery snowy road

**Figure 2:** Level 2 lane keeping feature oversteered car during skidding in snow on highway.

filling this void of literature, this paper's main contributions also include:

- Holistic analyses of the influences on ADS sensors directly induced or circumstantially brought by weather are presented statistically.

- Sensor fusion and mechanical solutions, perception enhancement algorithms, and pertinent technologies against weather conundrum are thoroughly reported. In the meantime, a quick index access to the corresponding literature is provided.

- Experimental validations of several solutions for perception enhancement under adverse weather are conducted.

- Perspectives of trends and future research directions regarding adverse weather conditions are proposed. Also the limitations that autonomous driving society could be facing are discussed.

The remainder of this paper is written in eleven sections: Section 2 is an overview of autonomous driving in general, while Section 3 presents the challenges and influences that weather brings to ADS sensors. Section 4 introduces sensor fusion and mechanical solutions related to weather; Section 5 presents perception enhancement methods experimental validation results with regard to each kind of weather; Sections 6, 7 & 8 state classification and assessment, localization, and planning & control. Auxiliary approaches in adverse weather solutions are given in Section 9. Section 10 collects the datasets, simulators and experimental facilities that support weather conditions. Section 11 provides analysis of trends / limits / future directions. Section 12 summarizes and concludes this work.

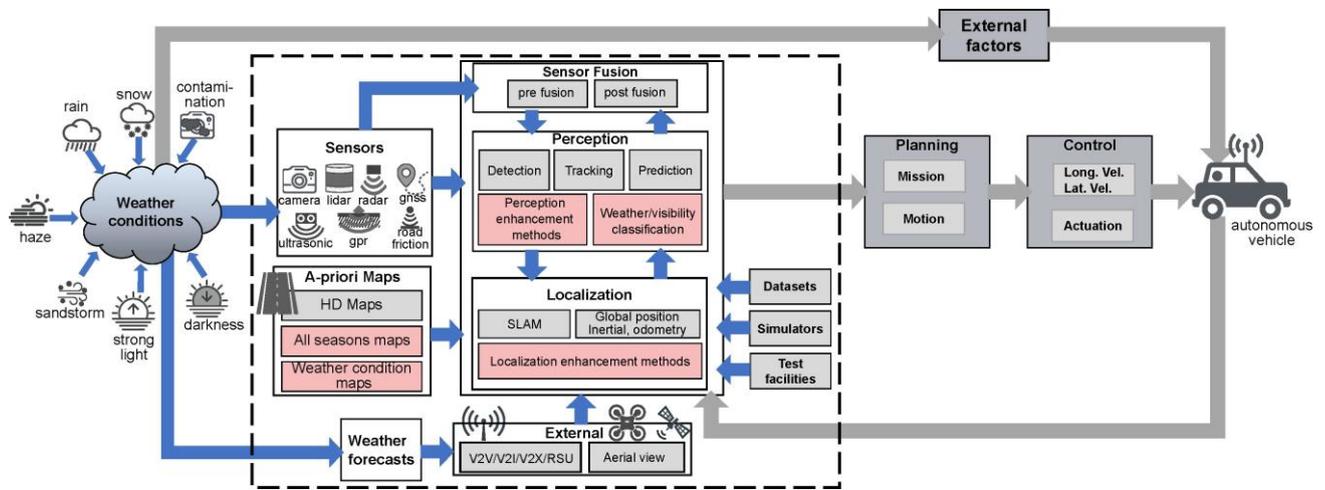

**Figure 3:** An architecture for self-driving vehicles agnostic to adverse weather conditions. Red blocks denote weather-related modules. Blue arrows denote the relationships among weather and perception and sensing modules. Gray arrows denote the relationships among ADS modules including external weather factors such as wind and wet road surfaces. This survey mainly focuses on the area enclosed in the dashed rectangle.

## 2. Overview of Autonomous Vehicles

An autonomous vehicle, also known as a self-driving car, is a vehicle that has the ability to sense its surrounding environment and navigate safely with little or no human input [21]. An information flow diagram shows the autonomous driving architecture and its relationship with weather conditions in Fig.3. As the vehicle sensors' information source, environmental states are directly affected by weather conditions. The changes increase difficulties for ADS to complete object detection, tracking and localization tasks out of impaired data, so planning and control would be different from normal too. Weather could also affect the ego vehicle itself with secondary effects such as winds and road surface conditions. Consequential effects brought by the change of ego vehicle or surrounding vehicles' state contribute to the change of environmental states in return and form a cycle. Fig. 4 contains the perception and sensing sensors that are covered in this paper when adverse weather is present.

The following subsections will explain the main components of an AV.

### 2.1. Sensors

Weather conditions happen in nature spontaneously and affect the environmental states in which AVs present. The changes in environments create discrepancies in the perception of vehicles sensors, which are the main components of ADS. The followings are the major perceptive sensors used in AVs.

#### 2.1.1. LiDAR

LiDAR is the core perception sensor in the autonomous driving field. The use of 3D-LiDAR on cars hasn't exceeded much more than a decade and has already demonstrated its indispensability in ADAS and AVs with high measurement accuracy and illumination independent sensing capabilities [22]. This 3-D laser scanning technology has some key attributes: measurement range, measurement accuracy, point density, scan speed and configuration ability, wavelength, robustness to environmental changes, form factor, and cost [2]. Fig.5 shows some common 3D LiDARs used for AVs.

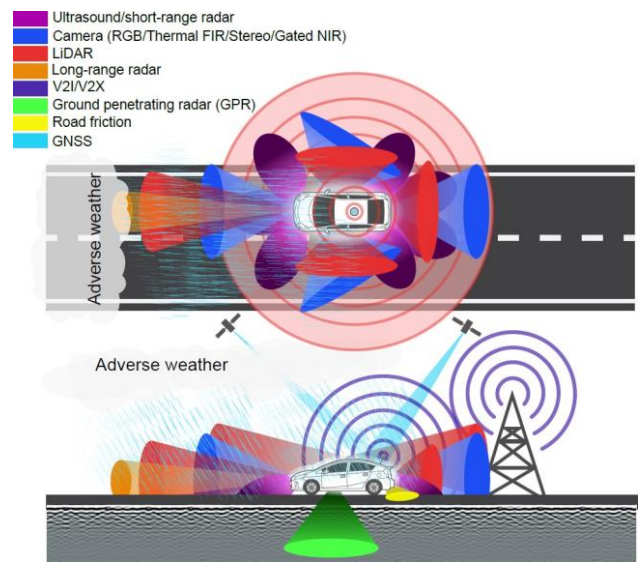

**Figure 4:** An illustration of sensors configuration in an autonomous vehicle, towards driving under adverse weather conditions. Sensor coverage denotes the general operating directions instead of real operating conditions, for reference only.

Modern LiDARs possess internal property flexibility. Many LiDAR models are equipped with the modality switch between the strongest signal return and the last signal return (also known as echo), and for clear weather, both give the same point cloud configuration. Suppose in a condition where fog is getting denser, the last return shows a larger overlap with the reference point cloud than the strongest return. LiDARs like the Velodyne HDL64-S3D [23] also provide the function of output laser power & noise ground level manual adjustment. While higher power output guarantees a longer detecting range, the right noise level choice can help improve accuracy, with the help of compatible denoising methods [24].

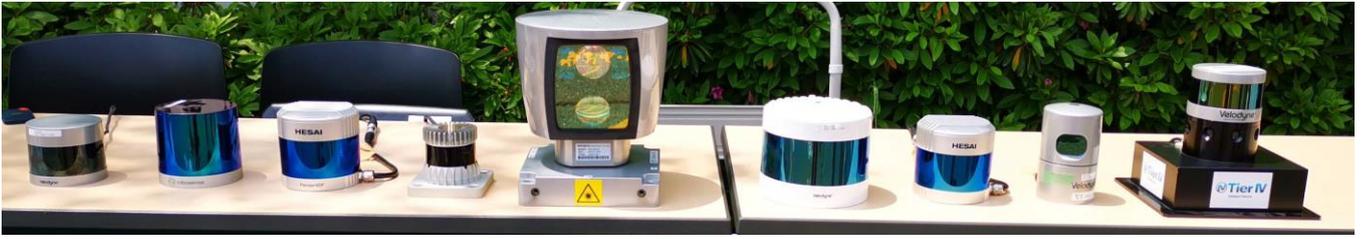

**Figure 5:** Multiple 3D LiDARs, from Velodyne, Hesai, RoboSense, Ouster, etc.

### 2.1.2. Camera

Although being a technology that is much older than LiDAR, camera is actually the one element that is absolutely not replaceable in ADS, while also one of the most vulnerable in adverse weather conditions. In a normal sense, the first reaction of people when putting the two words, car and camera, together is dashboard camera (dashcam). Adhered to the interior windshield, sometimes rear or other windows, dashcams continuously record the surroundings of a vehicle when plugged in with an angle as wide as 170° [25]. Numerous autonomous driving datasets started with dashcam recordings at an early stage while nowadays professional camera sets and fisheye lens cameras are being deployed to the data collecting task for an even larger field of view [26]. Cameras with specialties under particular situations such as night vision will be further discussed in sensor fusion in Sec.4.1.2 and potential future research in Sec.11.4.2.

### 2.1.3. Radar

The automotive radar system is a veteran in ADAS, long before LiDAR took the helm of autonomous driving. Automotive radar consists of a transmitter and a receiver. The transmitter sends out radio waves that hit an object (static or moving) and bounce back to the receiver, determining the object's distance, speed and direction. Automotive radar typically operates at bands between 24 GHz and 77 GHz which are known as mm-wave frequencies, while some on-chip radar also operates at 122 GHz. Radar can be used in the detection of objects and obstacles like in the parking assistance system, also in detecting positions, and speed relative to the leading vehicle like in the adaptive cruise control system [27].

There is also an FMCW form for radar where the frequency of the transmitted signal is continuously varied at a known rate which makes the difference between the transmitted and the reflected proportional to the time of flight, ergo range. Besides the speed measurement advantage, FMCW radar shows superior range resolution and accuracy in autonomous driving [28].

### 2.1.4. Ultrasonic

Among all the common automotive sensors, the one that is seldom being brought up in ADS modalities is the ultrasonic sensor. Being installed on the bumpers and all over the car body serving as the parking assisting sensor and blindspot monitor, ultrasonic has been the most diligent and cheapest sensor for a long time [29]. The principle of ultrasonic sensors is pretty similar to radar, both measuring the distance by calculating the travel time of the emitted electromagnetic wave, only ultrasonic operates at ultrasound band, around 40 to 70 kHz. In consequence, the detecting range of ultrasonic sensors normally doesn't exceed 11 m [30], and that restricts the application of ultrasonic sensors to close range purposes like backup parking. It's not like people didn't try to extend the effective range of ultrasonic and make it fit for long-range detecting [31], or ultrasonic has no use in autonomous driving. As a matter of fact, Tesla's "summon" feature uses ultrasonic to navigate through park space and garage doors [32].

### 2.1.5. GNSS/INS

Navigation or positioning systems are among the most basic technology found in robots, AVs, UAVs (Unmanned Aerial Vehicles), air crafts, marine vessels, and even smart-phones. Groves [33] provides a list of diverse measurement types and corresponding positioning methods.

The global navigation satellite system (GNSS) is an international system of multiple constellations of satellites, including systems such as GPS (United States), GLONASS (Russia), BeiDou (China), Galileo (European Union), and other constellations and positioning systems. GNSS operates in the L-Band (1 to 2 GHz) which can pass through clouds and rain, with a minimum impact on the transmitted signal in terms of path attenuation. GNSS sensors include one or more antennas, reconfigurable GNSS receivers, processors and memory. GNSS is often in combination with real-time kinematic positioning (RTK) systems using ground base-stations to transmit correction data.

Non-GNSS broadband radio signals are used for indoor, GNSS signal-deprived areas (i.e, tunnels), and urban positioning. Such systems include Wi-Fi based positioning systems (WPS), Bluetooth and Ultra-Wideband (UWB) beacons, landmarks, vehicle to infrastructure (V2I) stations, radio frequency ID (RFID) tags, etc.

Odometry and inertial navigation systems (INS) use dead reckoning to compute position, velocity and orientation without using external references. INS combines motion sensors (accelerometers), rotation sensors (gyroscopes), and also magnetic field sensors (magnetometers). For the advanced INS, fiber optic gyroscopes (FOG) are used: with no moving parts, and two laser beams propagating in opposite directions through very long fiber optic spools, the phase difference between the two beams is compared and it is proportional to the rate of rotation.

The combination of the above, such as GNSS with INS (GNSS+INS) and other sensors, with algorithms such as Kalman Filter and motion models, is a common approach to improve positioning accuracy and reduce drift. For example, the Spatial FOG Dual GNSS/INS of Advanced Navigation [34] has 8 mm horizontal position accuracy and about 0.005° roll/pitch accuracy.

### 2.2. Object detection & tracking and localization

Sensors can be treated as the means of ADS perception. Apart from perceiving the surrounding environments, the other main purpose of perception is to extract critical information that is essential to the safe navigation of an AV. In common sense, this information could mean cars and pedestrians on the road, traffic lights and road signs, and static objects like parked cars or city infrastructures, because those are what we would pay attention to when we are driving as humans. The ultimate reason behind the fact that we would pay attention to these things, also the number one rule of any kind of driving, is that we need to avoid colliding with them. In order to achieve this, we first need to know the existence of those things and where they are, i.e. object detection. General object detection in the computer vision area is to determine the presence of objects of certain classes in an image, and then determine the size of them through a rectangular bounding box, which is the label in nowadays autonomous driving datasets. YOLO (You Only Look Once) [35] evolves object detection into a regression problem with spatially separated bounding boxes and provides object class probabilities, which is now one of the most popular approaches. As for other sensors of ADS such as LiDAR and radar, the detection of an object is manifested by a signal return. However, given the LiDAR's relatively higher density, some methods such as PointPillars [36], Second [37] and Voxel-FPN [38], allow to correctly identify object classes in the point cloud.

Clearly, to avoid collisions, knowing only the location of others is not enough, but knowing the position of our own is equally important, and this is what localization is for. In terms of autonomous driving, localization is the awareness of an AV about its ego-position relative to a frame of reference in a given environment [39]. Some localization methods largely rely on the successful detection of certain elements in the surrounding environments, which we will talk about in Sec. 7.

But again, for an AV with high-level autonomy, merely the locations of itself and a detected object are inadequate. Only with a prospective trajectory of the other party can we make the right decision away from a collision in a relatively dynamic situation. The process of perceiving the heading direction and moving velocity and making the prediction of the opponent's trajectories is object tracking. As a part of the autonomous driving link, it's very important to the subsequent planning and control of an ego vehicle. This paper does not go deep into this for a better focus on weather conditions. More details on object detection & tracking can be found in [1].

### 2.3. Planning and Control

Planning normally means two subtasks in autonomous driving: global route planning, and local path planning. Global planner's job is to generate road options and choices when a given origin point and destination are assigned. Local path planning is about executing this road choice without hitting any obstacle or violating any rules, which is pretty much ensured by other modules of ADS like object detection. Details about general motion planning can be found in [40].

AV control is realized in three aspects: steering, throttle, and brake inputs, so is almost every vehicle [41]. The activation of certain actuators is a direct result of the decision made by the ADS mainframe based on the perceived information and processed analysis. The inputs to the steering wheel, throttle and brake control the direction, speed, and acceleration dynamic of an AV, which finally realize "driving". The vehicle can also implement low-level complementary safety measures at the electronic control unit (ECU), or vehicle control unit (VCU), by tapping into simple sensors such as sonar.

By bringing environmental state changes, weather conditions are affecting the normal perception functionalities of ADS sensors, which results in the intricacy in the following detection, planning, and control, etc. In the meantime, some other weather effects that are usually not picked up by perception sensors, such as wind and road surface condition changes, are also inflicting an ego vehicle directly and creating anomalies on the vehicle states. In return, the change in the ego vehicle's states further influences the normal operation of its sensors; the change in the ego vehicle's behaviors infects other surrounding vehicles and results in the change of the original environmental states since surrounding vehicles are part of the environments themselves. By now, a vicious cycle caused by the appearance of weather conditions is formed, as shown in Fig.3. That's what makes the research of autonomous driving in adverse weather conditions important.

## 3. Adverse Weather Influences

The weather challenge has been the bottleneck of ADS for a while now and numerous efforts have been done to

**Table 1**
The influence level of various weather conditions on sensors

| Modality | Light rain <4mm/hr | Heavy rain >25mm/hr | Dense smoke /Mist vis<0.1km | Fog vis<0.5km | Haze /Smog vis>2km | Snow | Strong light | Contamination (over emitter) | Operating Temperature (°C) | Installation complexity | Cost |
|---|---|---|---|---|---|---|---|---|---|---|---|
| LiDAR ($\lambda$ 850-950nm and 1550nm) | 2 | 3 | 5 | 4 | 1 | 5 | 2 | 3 | -20 +60 [42] | easy | high |
| Radar (24, 77 and 122 GHz) | 0 | 1 | 2 | 0 | 0 | 2 | 0 | 2 | -40 to +125 [43] | easy | medium |
| Ground-Penetrating Radar (100-400MHz) | 0 | 0 | 0 | 0 | 0 | 1 | 0 | 2 | -5 to +50 [44] | hardest | medium to high |
| Camera | 3 | 4 | 5 | 4 | 3 | 2 (dynamic) 3 (static) | 5 | 5 | -20 to +40 [45] | easiest | lowest |
| Stereo Camera | | | almost same as regular camera | | | | | | 0 to +45 [46] | easy | low |
| Gated NIR Camera [47] ($\lambda$ 800-950nm) | 2 | 3 | 2 | 1 | 0 | 2 | 4 | 3 | normally 0 to +65 [48] for InGaAs cameras | easy | low |
| Thermal FIR Camera ($\lambda$ 2-10$\mu$m) | 2 | 3 | 3 | 1 | 0 | 2 | 4 | 3 | -40 to +60 [49] | easy | low |
| Road-friction sensor* [50] (infrared) | 2 | 3 | 3 | 2 | 1 | 2 | 1 | 5 | -40 to +60 | medium | low |

```
The effect level each phenomenon causes to sensors:
0 - negligible: influences that can almost be ignored
1 - minor: influences that barely cause detection error
2 - slight: influences that cause small errors on special occasions
3 - moderate: influences that cause perception error up to 30% of the time
4 - serious: influences that cause perception error more than 30% but lower than 50% of the time
5 - severe: noise or blockage that cause false detection or detection failure
*Road-friction sensor operating relative humidity is < 95% but is able to measure 0~100% humidity
```

solve it. Actually the meteorology society has been studying adverse weather and the link to road safety for a very long time. Back when autonomous vehicles hadn't drawn much attention of the market, Perry et al. [51] thoroughly reviewed the hazards of slippery road surfacec brought by wet and ice, the visibility drop on the highway and the influences on the drivers' decision-making process, accidents induced by weather, and international efforts in promoting road safety. Time passes by and now AVs continue the battle.

Over a decade ago, R.H. Rasshofer et al. [52] had already attempted to analyze the influences of weather on the automotive laser radar system. They proposed a method previous to real tests and artificial environment—synthetic target simulation, which is to reproduce the optical return signals measured by reference laser radar under real weather conditions. Signal parameters like pulse shape, wavelength, and power levels were replicated and the influences of weather were presented in an analytical way. However, such an approach is no longer sufficiently reliable for ADS considering the real world is not invariant, and synthetic targets are impossible to reach exhaustivity. Still, they were one of the pioneers who started to confront adverse weather.

In order to better demonstrate the influences of some major weather conditions on ADS sensors, a detailed comparison is given in Table 1. It is worth noticing that level 3, moderate influences, that cause perception error up to 30% of the time in this table, could also mean up to 30% of the LiDAR point cloud is affected by noise, or up to 30% of the pixels in the camera images are affected by distortion or obscure. The same applies to level 4 influences, serious, as well.

### 3.1. Influence on LiDAR

Some key factors, like measurement range, measurement accuracy, and point density, could be interfered with by weather conditions and thus influence the normal operating of AVs. People have done tests and validations on LiDAR or the whole AV modality in adverse weather conditions ever since the concept appears, either in artificial environments like fog chambers, or real-world like Scandinavian snowfields, or even simulation environments.

#### 3.1.1. Rain and fog

For the most common weather, rain, when it's not extreme like a normal rainy day, it doesn't affect LiDARs and AV itself that much according to the research of Fersch et al. [54] on small aperture LiDAR sensors. The power attenuation due to scattering by direct interaction between laser beam and raindrops of comparable is almost negligible: the percentage diminution caused by rain at the criteria of how much signal stays above 90% of the original power is at the scale of two decimal spaces, and even for a more stringent criterion (99.5%) a loss of more than 10% of signal power has shown to be very unlikely. The effect from the wetting of the emitter window varies based on drop size, from max attenuation around 50% when the water drops are relatively small, to a minimal of 25% when the drop is about half the aperture size. It seems like the direct impact of rain which is wetting doesn't really shake the LiDAR but it's still worth noticing that when the atmosphere temperature is just below the dew point, the condensed water drops on the emitter might just be smaller than the lowest drop size in [54] and a signal with a power loss over 50% can hardly be considered a reliable one. Additionally, the influence of rain on LiDAR may not merely lie in signal power level but the accuracy and integrity of the point cloud could also be impacted which is hard to tell from a mathematical model or simulation.

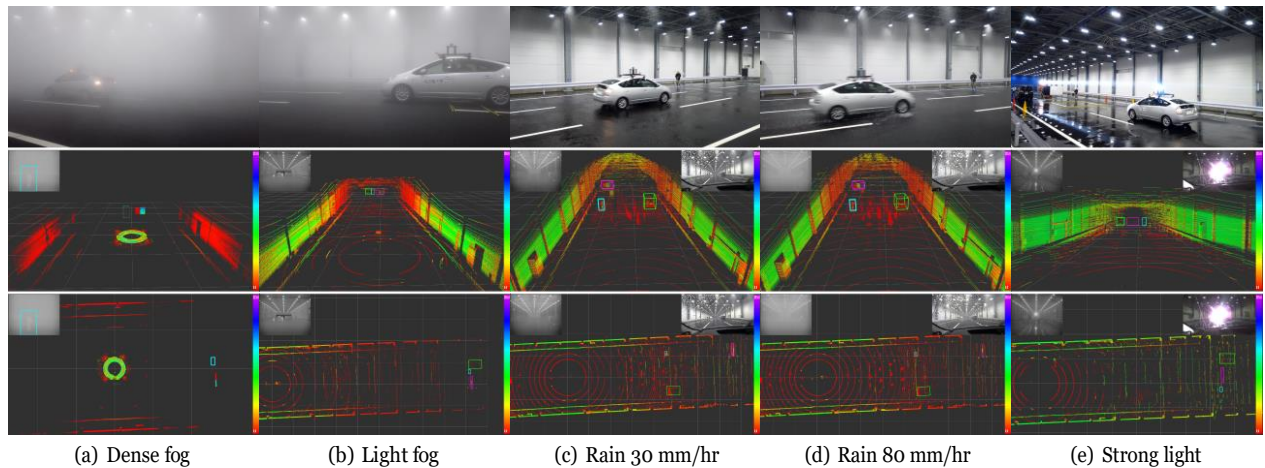

(a) Dense fog  (b) Light fog  (c) Rain 30 mm/hr  (d) Rain 80 mm/hr  (e) Strong light

**Figure 6:** Adverse weather results, top row depicts sample conditions, middle and bottom rows show the 3D LiDAR pointcloud, thermal camera image and RGB camera image (not available for fog experiments), targets of interest (human/mannequin, car and reflective targets) are highlighted. (a) dense fog with visibility of 17 m, (b) light fog with visibility 162 m, (c) rain fall setting of 30 mm/hr and average humidity of 89.5 %, (d) rain fall setting of 80 mm/hr and average humidity of 93 %, and (e) strong light at 200 klx at 155 A. Rain fall and visibility measurements using a VAISALA PWD12 laser disdrometer [53] at 875 $nm$, humidity was measured at 4 different stations, strong light used a 6000 W source with a color temperature of 6000 K and maximum current of 155 A.

More serious harm from rain happens when it becomes heavy or unbridled. Rains with a high and non-uniform precipitation rate would most likely form lumps of agglomerate fog and create fake obstacles to the LiDARs. As a result, we treat heavy rain the same as dense fog or dense smoke when measuring their effects. Hasirlioglu et al. [55] proved that the signal reflection intensity drops significantly at a rain rate of 40 mm/hr and 95 mm/hr by using the method of dividing the signal transmission path into layers in simulation and validating the model with a laser range finder in a hand-made rain simulator. Considering a precipitation rate of more than 50 mm/hr counts as violent rain and happens pretty rare even for tropical areas [56], the referential value here is relatively low in real life. Tests with real commercial LiDARs give a more direct illustration.

We can see from the LIBRE Dataset conducted by Carballo et al. [2] [59] that the point clouds of LiDARs in Fig.6 show discouraging results due to fog, rain and wet conditions. In the fog test, the highlighted human presence is only detectable by the LiDAR 13 m ahead on the dense setting but very few points to attempt recognition, and from 47 m ahead in the less dense setting. In the rain test, the highlighted objects were detected 24 m from the LiDAR, the difference is the level of noise due to the different rain settings. The artificial rain generated in a fog chamber, the Japan Automobile Research Institute (JARI) weather experimental facilities as shown in the first row of Fig.6 in this case, raised a new problem that most LiDARs detect the water comes out of the sprinklers as falling vertical cylinders which muddle the point cloud even more as illustrated in the third row of Fig.6(c) and 6(d). Fog chambers have come a long way from over a decade ago when researchers were still trying to stabilize the visibility control for a better test environment [60]. However, the real weather test might not completely be able to be replaced by fog chambers until a better replication system is available or the efficiency of such research might suffer consequently. We include an extensive review of weather facilities in Section 10.2.

*3.1.2. Snow*

Different from rain, snow is consisted of solid objects, snowflakes, and could easily shape themselves into much larger solid objects and become obstacles that either cause false detection of LiDAR or block the line of sight for useful detection. Very few tests on snow effects have been done given the fact that a snow test ground, like the fog chamber, is less easy to access, and the apparent danger of driving in snow.

Jokela et al. [57] tested AVs LiDAR performance in Finland and Sweden's snow conditions, mainly focused on the snow swirl caused by a leading car. Fig.7(a) shows the point cloud of accumulating multiple 3D scans as the ego vehicle moved behind the preceding vehicle. For an Ouster OS1-64 LiDAR, apart from some noise points near the sensor, the turbulent snow caused by the leading car and the ego car itself creates voids in the front and back view in the point cloud. It's worth noticing that this kind of point cloud is acquired in a condition where there is considerable accumulated snow on the ground with an interacting vehicle around, which is not very common in normal urban traffic

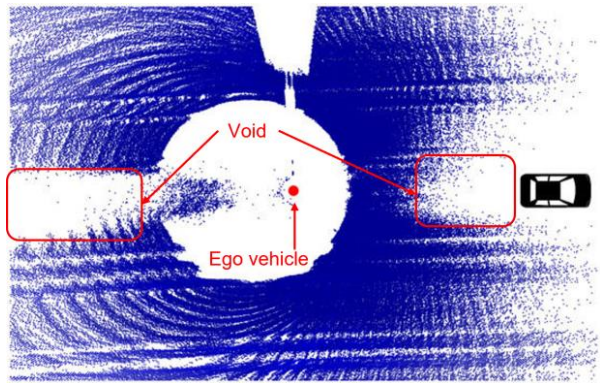

(a) Ouster OS1-64 point cloud in snow swirl. A few points of powder snow around the ego vehicle; missing view on both forward and back due to the turbulent snow caused by the leading vehicle and ego vehicle itself [57].

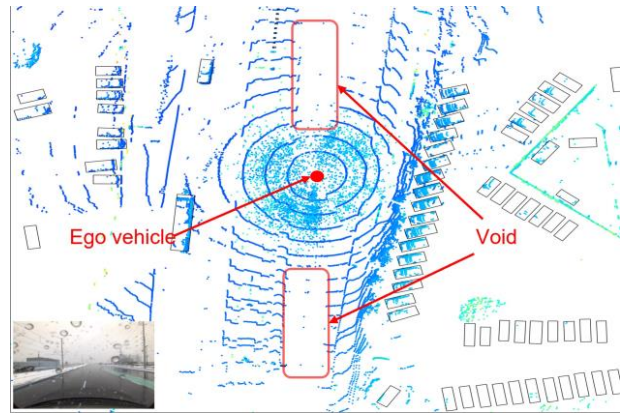

(b) Snow swirl effect without a leading vehicle. Black rectangles correspond to surrounding vehicles. Similar voids can be observed both in front of and behind the ego vehicle. The falling snow is sensed as similar as dense fog cloud around the ego vehicle. Point cloud scene from CADC Dataset [58].

**Figure 7:** LiDAR point clouds with swirl effect in snow weather. Image (a) courtesy of Dr. Maria Jokela, VTT Technical Research Centre of Finland Ltd.

and that's what makes it unique. We the authors explored the Canadian adverse driving conditions dataset [58] and tried to identify a similar swirl effect on paved road with no interacting vehicle ahead. It turned out that although the exact same result as Fig.7(a) is hard to capture, similar voids both in front of and behind the ego vehicle can also be observed, as shown in Fig.7(b). The voids can be caused by the swirl effect in heavy snow falls, as in Jokela et al. [57] findings, or due to accumulation of snow flakes on the optical window, or melted snow as water drops like shown in the bottom left inset of Fig.7(b). In a word, it's safe to say that snow swirl in the atmosphere or whirled from the ground could cause anomalies in LiDAR's point cloud and shorten the view distance.

One other factor in snow condition, low temperature, is also of concern. A LiDAR like Velodyne VLP-16 which was also used in [57], whose designed lowest operating temperature is $-10°C$, might not even stand a chance in a colder environment which is not that rare in the northern hemisphere. When the temperature change is at a large scale, such as from an extremely cold ($-20°C$) to an extremely hot ($+60°C$) environment, the time delay of LiDAR measurement would increase about 6.8 ns, which widens the LiDAR ranging by over a 1 meter and lower the precision at near field [61], not to mention the sensibility of photodetectors and range measurement.

### 3.1.3. Others

There are more weather phenomena that cause problems to transportation based on our common sense, such as sandstorms and smog. As rare as they might appear, they could be more serious problems than rain and snow for some regions like the Middle East or desert areas. However, due to the low attention and slow development under conditions when not even humans can drive, testing and researching in such weather can hardly be found. The part where LiDAR is involved in sandstorms or haze-smog weather is beyond the road—airborne or space. The CALIPSO high spectral resolution LiDAR [62] is used in satellites to monitor the Earth's atmosphere and can look through haze and sandstorms. Single-photon LiDARs are also frequently used in airborne LiDARs for 3D terrain mapping. Although such technology normally serves meteorology and oceanology [63], there is already single-photon avalanche diode (SPAD) LiDAR that has been used in automotive applications [64] due to its advantages in long-range capabilities (kilometers), excellent depth resolution (centimeters), and use of low-power (eye-safe) laser sources [65]. Future aerial LiDARs and UAVs are facing additional weather challenges as the situations in the sky is not quite the same as on the ground. Atmosphere turbulence can produce wind-affected and time-varying refractive gradients which lead to scintillation, beam spreading and wander [66]. The particular effect of such adversarial conditions on the autonomous driving area hasn't been studied in a methodically way as aerial LiDARs and UAVs haven't come to practical use yet, but it's safe to assume that they are going to need to overcome this problem to be able to serve the intelligent transportation system under hazy and turbulent conditions in the future. Details about aerial LiDARs and UAVs will be introduced in Sec.9.

### 3.2. Influence on radar

Radar seems to be more resilient in weather conditions. By examining the electromagnetic power attenuation in different rain rates [67] [68] from Fig.8, we can observe that the attenuation for radar at 77 GHz is at the level of 10 dB/km in a 25 mm/hr heavy rain, while 905 nm LiDAR's attenuation is about 35 dB/km under the same visibility below 0.5 km condition [69] [70]. According to Sharma and Sergeyev's simulation on non-coherent photonics radar which possesses lower atmosphere fluctuation, the detection range of the configuration of a linear frequency-modulated 77 GHz and 1550 nm continuous-wave laser could reach 260 m in heavy fog, 460 m in mild fog and over 600 m in heavy rain with SNR threshold at 20 dB [71]. Norouzian et al. [72] also tested radar's signal attenuation in snowfall. A higher snow rate yields larger attenuation is as expected, and wet snow shows higher attenuation because of the higher water absorption and larger snowflakes. Considering a snowfall with 10 mm/hr already has quite low visibility (< 0.1 km) [73], we yield that the specific attenuation for a 77 GHz radar in a 10 mm/hr snow is about 6 dB/km, which is seemly acceptable given the rain data.

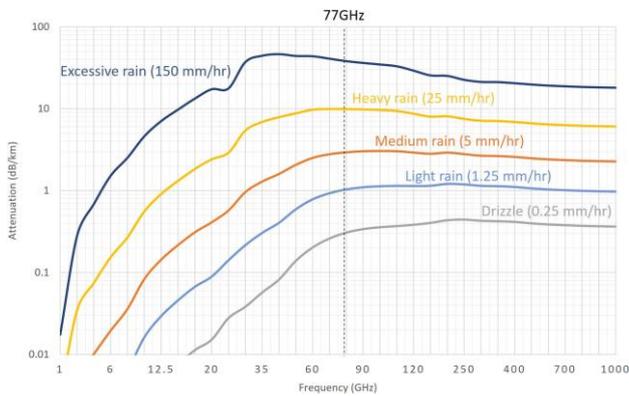

**Figure 8**: Electromagnetic power attenuation vs frequency in different rain rates.

In the research of Zang et al. [18], the rain attenuation and back-scatter effects to mm-wave radar and the receiver noise were mathematically analyzed. They conducted simulations including 4 different scenarios with radar detecting cars or pedestrians under different levels of rain rate. Results show that the back-scatter effect leads to the degradation of signal-to-interference-plus-noise ratio when the radar cross-section area is small like a pedestrian in heavy rain conditions. However, the degradation is at the single-digit level at a 100 mm/hr rain rate and their simulation expands the test variables up to a 400 mm/hr rain rate which is basically unrealistic in real-world because even if such an enormously high rain rate occurs, the condition of driving would be highly difficult.

No doubt that radar is objectively better adaptive to wet weather, but when compared with LiDAR, radar often receives criticisms for the insufficient ability in pedestrian detection and object shape & size information classification due to low spatial resolution. Akita et al. [74] have improved this by implementing long short term memory (LSTM) which can treat time-series data. What's more, one of the sensors used by the radar extension of Oxford RobotCar dataset [75] is a Navtech Radar CTS350-X 360° FMCW scanning radar [76] which possesses a measurement range up to 100-200 m and can handle Simultaneous Localization and Mapping (SLAM) solely in the dark night, dense fog and heavy snow conditions [77]. So the usefulness of radar has much more potential.

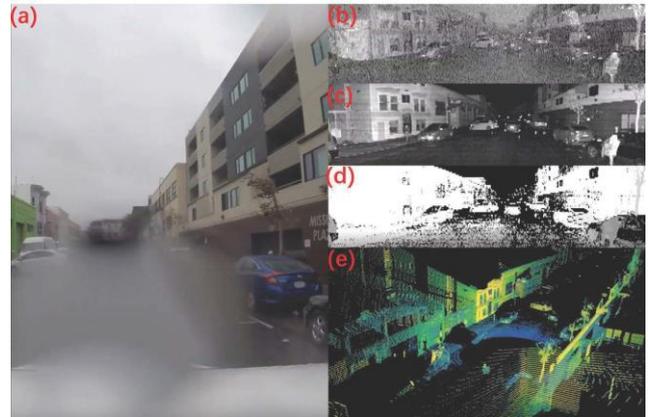

**Figure 9**: Camera vs LiDAR in rain condition [78]. (a) camera perspective; (b)intensity; (c) reflectivity; (d) noise; (e) 3D colored by intensity. Image courtesy of Ms. Kim Xie, Ouster Inc.

### 3.3. Influence on camera
#### 3.3.1. Rain and fog

A camera in rain, regardless of how high resolution, can be easily incapacitated by a single water drop on the emitter or lens [78], as shown in Fig.9. The blockage and distortion in the image would instantly make ADS lose the sense of the input data and fail to process correctly. As for fog, based on its density, it creates near-homogeneous blockages at a certain level which is a direct deprivation of information to cameras. Reway et al. [79] proposed a Camera-in-the-Loop method to evaluate the performance of the object detecting algorithm under different weather conditions. The environment model data are acquired by a set of cameras and processed by an object classification algorithm, the result is then fed to the decision maker which re-engages in the simulation environment and completes a closed loop. The result of up to 40% rise in miss rate in the night or fog pretty much fits with a common expectation of camera and proves that camera alone definitely cannot beat the weather.

#### 3.3.2. Snow

Winter weather like snow could affect the camera in one similar way like rain does when the snowflakes touch the lens or the camera's optical window and melt into ice-slurry immediately. What's worse, those ice water mixtures might freeze up again quickly in low temperatures and form an opaque blockage on the camera's line of sight.

Heavy snow or hail could fluctuate the image intensity and obscure the edges of the pattern of a certain object in the image or video which leads to detection failure [18]. Besides the dynamic influence, snow can extend itself to a static weather phenomenon. Most of the time the problems caused to transportation by snow are not instant because snow can accumulate on the surface of the earth and block road marks or lane lines, and might even form ice on the road [80]. That makes both humans and AVs drive under indeterminacy. Especially for cameras, the acquisition of data sources is compromised and the process of the whole autonomous driving is interrupted at the very beginning.

### 3.3.3. Light conditions

A particular weather phenomenon, strong light, directly from the sun or artificial light source like light pollution from a skyscraper may also cause severe trouble to cameras. Even LiDAR suffers from strong light in extreme conditions [2], showing a large area of black around the light source. As shown in Fig.6(e) upper right insets, too high an illumination can degrade the visibility of a camera down to almost zero and glares reflected by all kinds of glossy surfaces could make the camera exposure selection a difficult task [81]. HDR camera specializes in tough light conditions which will be introduced in Sec.11.4.2. With such technology applied to AVs, visibility drop due to sudden changes in light conditions like the entry and exit of a tunnel is largely mitigated. Benefiting from better color preservation, AV navigating performance when driving into direct sunlight can also be improved [82].

Another correlative issue caused by light is the reflection off reflective surfaces. If the reflection happens to be ideal, it might confuse the camera into believing it and transmitting a false signal. The lack of stereoscopic consciousness is the Achilles' heel of a normal camera. Human drivers frequently experience the problem of reflections on the windshield confusing the view on a very shiny day or when at night and there's illumination inside the cabin, which is exactly the reason why cabin light is not recommended when cruising at night time. The very same problem applies to cameras set up behind the windshields as well. Sometimes the reflections are an inferior mirage due to high road surface temperatures, sometimes are the mirror images of the car's interiors. This false information can hardly be identified in a 2D sense, and light conditions are never 100% friendly to cameras. Without the help of other sensors with the sense of depth in three-dimension, a camera couldn't take up the responsibility on its own. From another angle, the ineffectiveness of cameras here means the redundancy of ADS in weather conditions is impaired at a big level, so we need to do even better than just overcoming adverse weather to ensure total safety.

### 3.4. Miscellaneous issues

Apart from the normal weather that people are familiar with and could anticipate, there are always unexpected weather or phenomena caused by weather that could catch us off guard and that's also something that an AV needs to be prepared for. Like the flying stones hitting a car windshield, the casing of an ADS device like a LiDAR emitter window could also suffer direct strikes from a flying object, stones or sand blown up by strong winds, large pieces of hailstones in severe convective weather, and debris in general. A crack on the glass could affect the normal data receiving of signals or break the original images, also could create chances for air or moisture infiltration which might hinder electronic devices, and the worst scenario leads to fraction growth and finally fracture [83]. As one of the famous unpredictable phenomena, wind can bring more trouble. Contamination from unknown substances in the surroundings like blown up leaves and garbage could block the view of ADS sensors in a sudden. Particles from road dirt attached to the outer surface of the emitter window could worsen the LiDAR signal attenuation [84]. Tests with near-homogeneous dust particles being distributed on the surface of a scanner show a 75% reduction in LiDAR maximum range [85]. Although some of these scenarios can be considered rare events, it's better to be prepared because ADS safety cannot afford any tolerance.

Ultrasonic is among the sensors that are hardly considered in the evaluation of weather influences, but it does show some special features. The speed of sound traveling in air is affected by air pressure, humidity, and temperature [86]. The fluctuation of accuracy caused by this is a concern to autonomous driving unless enlisting the help of algorithms that can adjust the readings according to the ambient environment which generates extra costs. Nonetheless, to keep an open mind in ADS modalities, ultrasonic does have its strengths, considering its basic function is hardly affected by harsh weather. The return signal of an ultrasonic wave does not get decreased due to the target's dark color or low reflectivity, so it's more reliable in low visibility environment where cameras may struggle, like high-glare or shaded areas beneath an overpass. Additionally, the close proximity specialty of ultrasonic can be used to classify the condition of the road surface. Asphalt, grass, gravel or dirt road can be distinguished from their back-scattered ultrasonic signals [87], so it's not hard to imagine that the snow, ice or slurry on the road can be identified and help AV weather classification as well. Bottom line is, ultrasonic sensors provide another layer of redundancy when other sensors like LiDAR and cameras are crippled in some abominable environments and at least keep the AV from collisions while maintaining its original duties.

Now we know that weather conditions directly affect the environmental states and impair ADS sensors' ability to perceive; or affect the vehicles' states with secondary effects such as winds and road surfaces. There are still consequential effects brought by the change of ego vehicle or the surrounding vehicles' state. Walz et al. [88] benchmarked the spray effect on cameras and LiDARs caused by the nearby cutting vehicles under both day and night conditions. They proved that the spray would cause "ghost" targets for Velodyne HDL64-S3D LiDAR and severe blockage for the non-rotating Velodyne VLP32C LiDAR that doesn't have the centrifugal forces to throw away the water droplets or any other self-cleaning mechanism; and also false positives around the rear lights, wrong object dimensions, and missing detections for camera's case. Vargas Rivero [89] used synthetic data to construct a virtual scene where an obstacle in a water spray region has the same detection characteristics as a real object in terms of intensity, echo number and occlusion. They generated augmented data from the detection of this virtual scene and obtained good LiDAR point cloud data for classifier training under water spray conditions from a leading vehicle. This effect sometimes even outdoes the direct effect from weather as the spray plume from the leading vehicle is often violent and erratic, just like the snow swirl in [57].

**Table 2**
Sensor fusion and mechanical solutions

| Sensor Fusion | | | Mechanical Solutions | | | | | |
|---|---|---|---|---|---|---|---|---|
| Radar dominant | Thermal camera dominant | LiDAR+Radar +Camera+Auxiliary | Sensor housing and wipers | Heater | Hydrophobic coating | Lower reflection | Windshield protection | Sensor recalibration |
| [92] | [94] | [97] | [100] [101] [102] | [103] [104] | [105] [106] | [107] | [108] [83] | [109] [110] |
| [93] | [95] | [81] | | | | | | |
| | [96] | [98] | | | | | | |
| | [47] | [99] | | | | | | |

Signals from satellite-based navigation systems, such as GPS, Galileo and others, experience some attenuation and reflection with passing through water in the atmosphere and other water bodies. As analyzed by Gernot [90], water is a dielectric medium and a conductor. Electromagnetic waves will experience attenuation due to the rotation of water molecules according to the electric field which causes energy dissipation. Also, moving charges in the water body will reflect and refract the wave, and this happens at the air-water and water-air interfaces.

Balasubramaniam and Ruf [91] studied the effects of rain and winds on GNSS reflectometry (GNSS-R), their model considers path attenuation, modified surface roughness and downdraft winds. Their study finds very little attenuation in the L-Band due to raindrops, with about 96% transmissivity for mild rain below 30 mm/hr. While the L-Band is able to penetrate even heavy rain, the effect of wind tends to increase attenuation.

In Gernot's work [90], GPS signal experienced a significant loss over 9.4 dB when passing through a 1 mm layer of liquid water, but only 0.9 dB when passing through a 4 cm layer of snow, and 1.7 dB for a 14 cm layer of snow. This experiment suggests that bodies of water in the form of wet roads and puddles will further affect the signal-to-noise ratio of GNSS.

## 4. Sensor Fusion and Mechanical Solutions

The serious influences that weather causes on autonomous driving encourage people to work on solutions. Before the wild spread and use of machine learning and AI training techniques, the more straightforward solutions in the eyes of most of the manufacturers and Auto factories are sensor fusion modalities agnostic to weather, and mechanical components that help mitigate the effects of weather. Table 2 shows the literature and newsletters coverage with regard to different types of sensor fusions and mechanical solutions.

### 4.1. Sensor fusion modalities

By now, it's almost well established that the traditional LiDAR or Camera architecture alone is not going to navigate through adverse weather conditions with enough safety assurance. But two forces combining together would be a different story with the additional strength, just as the enhanced point cloud from a LiDAR + Camera fusion shown in Fig.10. [111] pointed out that a sensor fusion modality outdoes every sensor on their own including LiDAR, camera and radar, not only in weather conditions but also the overall perception performance. As a result, groups from all over the world come up with their own permutation and combination with the addition of radar, infrared camera, gated camera, stereo camera, weather stations and other weather-related sensors, like the one shown in Fig.11. And of course, fusion modalities need calibrations to ensure the synchronization of all the participated sensors for the best performance. An excellent example with centimeter-level accuracy is the multi-sensor fusion toolbox developed by Monrroy Cano et al. [110] whose framework contains both LiDAR-to-LiDAR and Camera-LiDAR extrinsic algorithms that can ease the fusion of multiple point clouds and cameras with only one-time calibration. This toolbox is integrated into the open source autonomous driving framework known as Autoware [112].

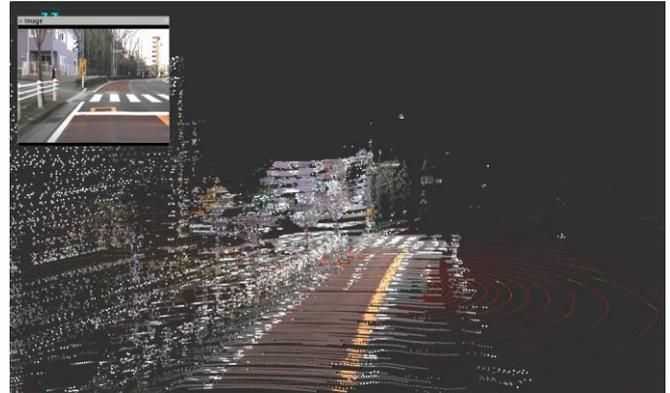

**Figure 10:** Point cloud of a LiDAR + Camera Fusion. Point cloud colored by pixels data fusion.

#### 4.1.1. Radar dominant

Yang et al. [92] brought up a modality called RadarNet, which exploits both radar and LiDAR sensors for perception. Their early fusion exploits the geometric information by concatenating both LiDAR and radar's voxel representation together along the channel dimension, and the attention-based late fusion is designated to specifically extract the radar's radial velocity evidence. They validated their modality in the nuScenes dataset [113]. Although no specific tests in adverse weather conditions were conducted, we know that the nuScenes dataset unbiasedly collected rain conditions in Boston and Singapore, so the robustness of such classic fusion is proven, especially in the improvement of long-distance object detection and velocity estimation.

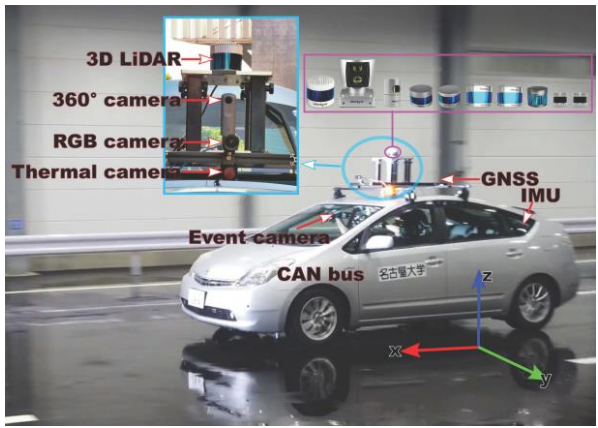

**Figure 11:** The Toyota Prius used for ADS tests from Nagoya University. The LiDAR sensor, alongside other sensors, is bolted on a plate which is mounted firmly on top of the car.

Liu et al. [93] raised a robust target recognition and tracking method combining radar and camera information under severe weather conditions, with radar being the main hardware and camera the auxiliary. They tested their scheme in rain and fog including night conditions when visibility was the worst. Results show that radar has pretty high accuracy in detecting moving targets in wet weather, while the camera is better at categorizing targets and the combination beats LiDAR alone detection by over a third. Their radar also shows good stability in tracking vertical targets but not horizontal targets due to the limited field of view (FOV). Radar and camera together reach close to the LiDAR tracking ability and they concluded that this mixture stands a good chance in adverse weather conditions.

Fritsche et al. [114] used a 2D high bandwidth scanner, the mechanical pivoting radar (MPR) [115], to fuse with LiDAR data to achieve SLAM in a low visibility fog environment. The MPR only has a 15 m measurement range but the one ability to penetrate fog is more than enough to prove itself useful in landmark searching and make up for what the LiDAR is missing. This fusion was tested on a robot instead of an AV.

### 4.1.2. Camera dominant

FLIR System Inc. [94] and the VSI Labs [116] tested the world's first fused automated emergency braking (AEB) sensor suite in 2020, equipped with a thermal long-wave infrared (LWIR) camera, a radar and a visible camera. LWIR covers the wavelength ranging from 8 $\mu$m to 14 $\mu$m and such camera operates under ambient temperature known as the uncooled thermal camera. This sensor suite was tested along with several cars with various AEB features employing radar and visible camera against daytime, nighttime and tunnel exit into sun glare. The comparison showed that although most AEB systems work fine in the daytime, normal AEB almost hit every mannequin under those adverse conditions, while the LWIR sensor suite never knocked down a single one. As a matter of fact, LWIR camera also exhibits superior performance in thick fog conditions when scattering loss is very high compared to MWIR (3 $\mu$m - 5 $\mu$m) and SWIR (0.85 $\mu$m - 2 $\mu$m) [117].

Vertens et al. [118] went around the troublesome nighttime images annotation and leveraged thermal images. They taught the network to adapt and align an existing RGB-dataset to the nighttime domain and completed multi-modal semantic segmentation. Spooren et al. [95] came up with a multi-spectral active gated imaging system that integrated RGB and NIR cameras for low-light-level and adverse weather conditions. They designed customized filters to achieve a parallel acquisition of both the standard RGB channels and an extra NIR channel. Their fused image is produced with the colors from the RGB image and the details from the NIR. John et al. [96] also proposed a visible and thermal camera deep sensor fusion framework that performs both semantic accurate forecasting as well as optimal semantic segmentation. These might be some of the most cost-effective solutions for weather conditions but particular gated CMOS imaging systems are still being developed [47].

It should be noted that even though thermal cameras can have better performance than regular cameras and can definitely be tested in winter, the operating temperatures provided by the manufacturers have certain lower bounds as shown in Table 1, which might seriously restrain the practical use of such sensors during cold winter even if it's a clear day. The durability of such temperature-sensitive devices needs further validation in real environments in the future to ensure their usefulness.

### 4.1.3. Comprehensive fusions

Kutila et al. [97] raised an architecture called the Robust-SENSE project. They integrated LiDAR with long (77GHz) & short (24GHz) range radar and stereo & thermal camera while connecting the LiDAR detection layer and performance assessment layer. That way, the data gathered by the supplementary sensors can be used in the vehicle control layer for reference when the LiDAR performance is assessed as degrading down to a critical level. They tested the architecture with a roadside LiDAR in a foggy airport and collected performance data while keeping the hardware components cost at a considerably low price (< 1000 Euros). Although the comparability with an AV test drive is not ideal, the concept of hardware and software complementation is one of the bases of AV weather adaptation. Radecki et al. [81] extensively summarized the performance of each sensor against all kinds of weather including wet conditions, day & night, cloudy, glare, and dust. They formulated a system with the ability of tracking and classification based on the probability of joint data association. Their vision detection algorithm is realized by using sensor subsets corresponding to various weather conditions with real-time joint probabilistic perception. The essence of such fusion is about real-time strategy shift. Sensor diversity improves the perception ability's general lower bound, but the intelligent choice of sensor weighting and accurately quantified parameters based on the particular weather determine the ceiling of the robustness and reliability of such modalities.

Bijelic et al. [98] from Mercedes-Benz AG present a large deep multimodal sensor fusion in unseen adverse weather. Their test vehicle is equipped with the following: a pair of stereo RGB cameras facing front; a near-infrared (NIR) gated camera whose adjustable delay capture of the flash laser pulse reduces the backscatter from particles in adverse weather [99]; a 77 GHz radar with 1° resolution; two Velodyne LiDARs namely HDL64 S3D and VLP32C; a far-infrared (FIR) thermal camera; a weather station with the ability to sense temperature, wind speed & direction, humidity, barometric pressure, and dew point; and a proprietary road-friction sensor whose purpose not specified, presumably for classification use. All the above are time-synchronized and ego-motion corrected with the help of the inertial measurement unit (IMU). They claimed their fusion entropy-steered, which means regions in the captures with low entropy can be attenuated, while entropy-rich regions can be amplified in the feature extraction. All the data collected by the exteroceptive sensors are concatenated for the entropy estimation process and the training was done by using clear weather only which demonstrated a strong adaptation. The fused detection performance was proven to be evidently improved than LiDAR or image only under fog conditions. The blemish in this modality is that the amount of sensors exceeds the normal expectation of an ADS system. More sensors require more power supply and connection channels which is a burden to the vehicle itself and proprietary weather sensors are not exactly cost-friendly. Even though such an algorithm is still real-time processed, given the bulk amount of data from multiple sensors, the response and reaction time becomes something that should be worried about.

### 4.2. Mechanical solutions
#### 4.2.1. Protection and cleaning

It's always the first instinct in a human's vein to solve a problem mechanically like tightening a screw. All the problems that weather has been causing to an AV most definitely caught the attention of the major automotive enterprises and efforts have been made to get around them with simple, low-cost mechanical solutions. For example, growing out of snowy Scandinavia, Volvo first noticed that the snow swirl caused by the leading car blocked the view of sensors and made them freeze up. After moving the sensors to several locations on a car ending up in failures, Volvo finally decided to nestle the radar and cameras behind the windshield and keep them from the hassle of snow permanently [100]. However, this is more of a compromise than a real solution considering the installation of these sensors inside the cabin does not really comply with the current market where human drivers still have the seat behind the wheel, and some sensors such as LWIR thermal cameras cannot be installed behind windows because their wavelength won't go through glasses.

Having experienced the importance of the windshield, it's natural for us to think of trying installing the same mechanism for ADS sensors, as in designated windshields. Waymo and Uber are among the proponents who cover the ADS sensors, mostly LiDAR and cameras, inside a shell housing and attach a small wiper on it [101] [102]. Similar to the windshield and wipers, now the sensors can enjoy almost the same benefits as inside the cabin. It's not only rain and frost we are talking about, but contamination such as bird droppings and bug splatters which are both common and pernicious to perception. Plus, a shell housing can at least protect the fragile sensors from blunt traumas which might come from flying stones, hails or corrosive substances like acid rain or snow melting agents. The only concern is that sensors behind the windshield or a shell casing on an AV require automatic wipers for raindrop and contamination cleaning. Optical or electrical rain sensors in current modern car automatic windshield wiper systems have the ability to fulfill this duty, but the technology for contamination detection is still on the way.

But of course, more mechanical parts mean more moving components and potential risks of dysfunction or damage. Some might prefer a mechanism that is static as the alternative to wipers, such as a heater. When driving in winter or cold rain, we tend to turn on the heater on the rear window and rear-view mirrors to get rid of the mist or frost to keep a clear view [103]. Hence, a heater can also certainly help keep a sensor emitter or a camera lens clean. State-of-art technologies apart from embedded heat-resistance wires have been developed to help realize this task. Canatu Company created carbon nanobots that can generate 10°C in less than 6 seconds, and are then deposited onto the plastic covers of sensors and headlamps [104]. Such a mechanism not only offers ice-free sensor surfaces, but also almost doesn't consume any energy which could be critical for electric cars (EVs).

There is a widely used process in industries such as jewelry and glasses which is ultrasonic cleaning [119]. This technology employs piezoelectronics (PZT) to generate a 35 kHz ultrasound and vibrate the emitter surfaces to transfer the dirt into a thin film of water or cleaning fluid. Then the fluid is atomized from the surface, taking away the dirt. This mechanism is a great arsenal for bug splatter, dirt, road debris, and most importantly, fog condensation.

#### 4.2.2. Passive accessories

Besides the proactive machinery, there are also passive solutions that help ease the adverse effects. Hydrophobic membrane has been applied to car windshields for a couple of decades now and is proven to be able to improve visual distance and decrease the minimum visual angle by almost 34% [120]. Research also shows that this improvement benefits the detection of objects and road sign reading greatly which are essential for ADS [105]. Therefore, there is no reason to reject hydrophobic coatings on the sensor covers of AVs, and PPG Inc. (Pittsburgh Plate Glass Company) has already developed relevant water repellent products on car glasses [106].

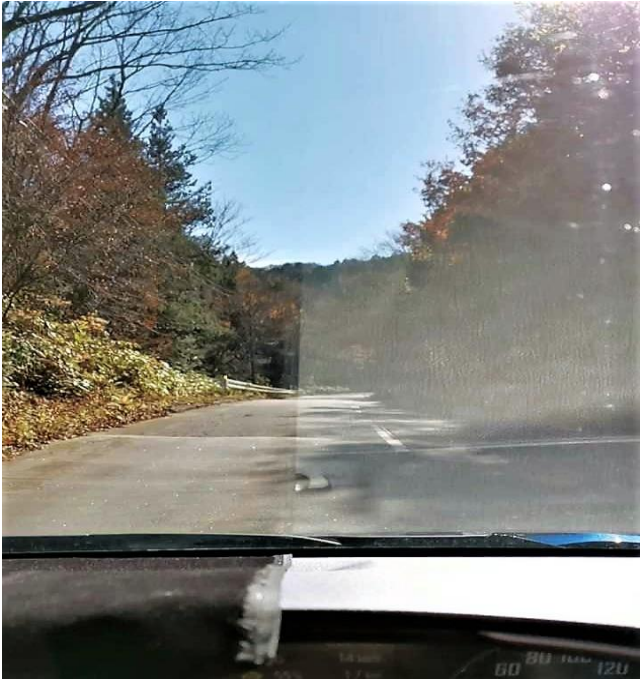

**Figure 12:** KoPro Musou Black visual light absorbing flock sheet on the left side of the dashboard alleviates reflections of car internals compared to the right side and improves visibility under sunlight.

Beyond all the advanced technologies, sometimes mechanical solutions can be as simple as putting a piece of cloth over the dashboard. A Japanese company called Ko-Pro develops extremely low reflection paint (called Musou Black) also manufactures a kind of flock sheet which is made of rayon-base fabric with a reflectance rate in visible light and NIR of less than 0.5% [107]. Simply placing the flock sheet on the dashboard of a car, very few reflections of the car's interior can be found on the windshield and the overall perception from within the cabin on sunny days is boosted so obviously even for a human eye, as shown in Fig.12. Such improvement in camera perception for the price of only about 30 US dollars might just be currently the best price-quality ratio in mechanical solutions.

### 4.2.3. Integrity and re-calibration

Despite all the additional mechanical parts, the unavoidable problem with shell housings remains the same: as the last protection layer between the outside world and the sensors, the shell itself needs an alert system as much as the windshield does. With some of the sensors like a dash-cam still being installed behind the windshield, even if there's no human in the front row seats, the structural integrity of the shield glasses still needs to be ensured. Even though nowadays's windshields are made with shatterproof glass with polyvinyl butyral embedded in them [108], it's still not impregnable. Apple car has already filed a patent using infrared light to detect chips and fractures on the windshield at an early stage to prevent crack extensions and splinters [83], which can be seen as a part of the road that Apple is paving to the AV security.

Mechanical components structural integrity pre-warning is one side of the coin, while the other side is the re-calibration. [121] shows that AV sensors suffer significant impact from lifetime effects and result in degradation. Current popular setups of ADS sensors are LiDARs on the roof rack; cameras all around the car body or nestled behind the windshield; radar and ultrasonic parts hidden inside the grille or front/back bumpers. And since each sensor works at a state with narrow tolerances, any impact or status change that happened to whatever mechanical parts could knock the sensors out of alignment easily. A slight change in the shell casing's curve or the general position could cause the cameras to lose focus; a deformation not visible to naked eyes in the support structures of LiDARs or radars could shift their field of view out of the designated interval; even the height change of the car's gravitational center due to weight or tire pressure change may cause the perception results of sensor different than before. Bosch uses green lasers that shoot visible straight lines to help do the calibration. The American Big Three (Ford, GM, Stellantis [previously FCA]) all require dynamic re-calibrations to their cruise and lane-keeping systems which includes test drivings in the Operational Design Domain (ODD) with speed requirements. Some models from Honda and Mercedes require both static and dynamic re-calibration on top of typical repairs which could easily cost a fortune [109]. Such points might not exactly fall within the category of weather, but adverse conditions definitely contribute to the sensors' lifetime effects. As a result, in order to remedy the rising cost of AV re-calibration, the mechanical parts of AVs need even higher durability and robustness.

## 5. Perception Enhancement Methods and Experimental Validations in Different Weather Conditions

As established in the previous context, the signal intensity attenuation and noise disturbance caused by weather phenomena impair the ADS sensors' abilities to carry out their original duties and make the risk index of autonomous driving climb rapidly. Despite the compensation of mechanical protections and advanced navigating techniques, the core of ADS, object detection, is the one that's worth the most attention and effort in fighting against adverse weather conditions. In this section, research and works regarding perception enhancement will be introduced, according to each kind of weather respectively. An index of literature covered in this paper regarding adverse weather solutions against each kind of weather type can be found in Table 3.

### 5.1. Rain

Rain is something that's considered predestined because it falls from the sky. Knowing rain is not avoidable and its influence on the ADS sensors, it's natural to want AV to realize the presence of rain for starters. In meteorology, rain is observed and measured by weather radar and stationary rain gauges. Considering carrying a weather station on a car like the Mercedes group [98] is not practical for commercial generalization, people started in an early stage to realize vehicle-based binary (wet/dry) precipitation observations when the purpose was not for autonomous driving yet [177] [178]. Although these successfully achieved precipitation perceiving in the real-time field, the data were pulled from simulation and had only been validated against weather radar but not real-world data. Karlsson et al. did an estimation on the real-time rainfall rate out of automotive LiDAR point cloud under both static and dynamic conditions in a weather chamber using probabilistic methods [127]. Bartos et al. [125] raised an idea of producing high-accuracy rainfall maps using windshield wipers measurement on connected vehicles in 2019. It's a very leading concept considering the network of connected vehicles has not been constructed on a large scale. Simply the status (on/off) of windshield wipers serves as the perfect indicator of binary rainfall state compared to traditional sensing methods like rain gauges. This work is supposed to help city flash flood warnings and facilitate stormwater infrastructure's real-time operation, but the involvement of cars provides a line of thought on vehicle-based rain sensing.

**Table 3**
Summary of solutions against different kinds of adverse weather types

| Rain | | Fog/Smoke/Haze | | Snow | | Light conditions related | | Contamination |
|---|---|---|---|---|---|---|---|---|
| Sensor solution | De-Raining | Perception | De-Hazing | Perception | De-snowing (snowfall) | Strong light /Reflectance | Shadow | Dirt/Dust /Soiling |
| Fusion [81] LiDAR [122] LiDAR [123] LiDAR [124] Others [125] LiDAR [126] LiDAR [127] | Camera [128] Camera [129] Camera [130] Camera [131] Camera [132] Camera [133] Camera [134] Camera [135] | LiDAR [136] LiDAR [137] LiDAR [97] LiDAR [24] Fusion [98] LiDAR [138] Fusion [139] LiDAR [140] LiDAR [141] Fusion [142] LiDAR [143] | Camera [144] Camera [145] Camera [146] Camera [147] Camera [148] Camera [149] Camera [150] Camera [151] Camera [152] Camera [153] Camera [154] Camera [155] Camera [156] | Camera [157] Fusion [158] Fusion [159] | LiDAR [160] Camera [161] Camera [162] LiDAR [163] | Camera [164] Camera [165] Camera [166] Camera [167] Camera [168] Camera [169] Camera [170] Camera [171] | Camera [172] Camera [95] Camera [173] Camera [174] | LiDAR [84] Camera [175] Camera [176] |

From the previous context, we know that drizzling and light rain barely affects main ADS sensors' performance like LiDAR, but we do fear when the rain rate rises. A. Filgueira et al. [126] thought of quantifying the rain's influences on LiDAR. They put a stationary LiDAR by the roadside and compared the range detection change, signal intensity change, and the number of detected points changes with regard to several detection areas including road signs, building facades and asphalt pavement. The problem is, not only their test scenario is a stationary one, but the fact that asphalt pavement is perpendicular to rain and the building facade is parallel to rain also affects the impartiality of quantifying standards. Still, their work initiated the idea of quantifying rain by directly quantifying the LiDAR performance change caused by rain. Along this train of thought, it's possible to set thresholds of rain effects when certain actions needed to be taken to counter the influences [179], as long as the benchmarks can be precise and inclusive. Goodin et al. [124] tried this task in a more specific way. They used only two parameters: rain rate, as manifested by the rain scattering coefficient, and the max range of the LiDAR sensor for a 90% reflective target in clear conditions, to successfully generate a quantitative equation between rain rate and sensor performance. Their design and validation were conducted under a simulation environment where rain rates are easily controlled. Even though no field validation has been done, combining proper precipitation sensing modalities, now it's totally possible to know when the LiDAR performance has degraded down to a critical level according to rain rate. This helps simplify the decision-making process a lot by setting quantifiable benchmarks.

Before new LiDAR technology emerges besides 1550 nm wavelength, camera is still the major focus of perception enhancement in rain, mainly in terms of de-raining technique, which has been deeply studied by the computer vision field. The detection and removal of raindrops can be divided into falling raindrops and adherent raindrops that accumulated on the protective covers of cameras [135]. For rain streaks removal, several training and learning methods have been put to use including Quasi-Sparsity-based training [130] and continual learning [131]. Quan et al. [133] proposed a cascaded network architecture to remove rain streaks and raindrops in a one-go while presenting their own real-world rain dataset. Their raindrop removal and rain streak removal work in a complementary way and the results are fused via an attention-based fusion module. They effectively achieved de-raining on various types of rain with the help of neural architecture search and their designated de-raining search space. Ni et al. [132] introduced a network that can realize both removal and rendering. They constructed a Rain Intensity Controlling Network (RIC-Net) that contains three sub-networks: background extraction, high-frequency rain streak elimination and main controlling. Histogram of oriented gradient (HOG) and auto-correlation loss are used to facilitate the orientation consistency and repress repetitive rain streaks. They trained the network all the way from drizzle to downpour rain and validation using real data shows superiority.

Like common de-noising methods, a close loop of both generation and removal can present better performance. H. Wang et al. [134] handled the single image rain removal (SIRR) task by first building a full Bayesian generative model for rainy images. The physical structure is constructed by parameters including direction, scale and thickness. The good part is that the generator can automatically generate diverse and non-repetitive training pairs so that efficiency is ensured. Similar rain generation is proposed by Ye et al. [129] using disentangled image translation to close the loop. Furthermore, Z. Yue et al. [128] surpassed image frames and achieved semi-supervised video de-raining with a dynamic rain generator. The dynamical generator consists of both an emission and transition model to simultaneously construct the rain streaks' spatial and dynamic parameters like the three mentioned above. They use deep neural networks (DNNs) for semi-supervised learning to help the generalization for real cases.

While de-raining has been extensively studied using various training and learning methods, most of the algorithms have met challenges on adherent raindrops and performed poorly when the rain rates or the dynamism of the scene get higher. Detection of adherent raindrops seems to be easy to achieve given the presumed optical conditions are met, but real-time removal of adherent raindrops inevitably brings the trade-off of processing latencies regardless of the performance. Peak signal-to-noise ratio (PSNR) and Structural Similarity (SSIM) metrics although widely implemented and faintly improved, don't seem exactly promising in the line of raindrops detection and removal compared to deep-learning and convolutional neural networks (CNN) [135].

### 5.2. Fog
#### 5.2.1. *Fog in point clouds*

Fog plays a heavy role in the line of perception enhancement in adverse weather conditions, mainly due to two reasons. First, the rapid and advanced development of fog chamber test environments, and second, the fog format commonality of all kinds of weather including wet weather and haze and dust, in other words, the diminution of visibility in a relatively uniform way. Early in 2014, Pfennigbauer [136] brought up the idea of online waveform processing of range-finding in fog and dense smoke conditions. Different from the traditional mechanism of time-of-flight (TOF) LiDAR, their RIEGL VZ-1000 laser identifies the targets by the signatures of reflection properties (reflectivity and directivity), size, shape and orientation with respect to the laser beam, which means, this echo-digitizing LiDAR system is capable of recording the waveform of the targets which makes it possible to identify the nature of the detected target, i.e. fog and dense smoke by recognizing their waveforms. Furthermore, since the rate of amplitude decay caused by the fog follows a certain mathematical pattern with regard to the density of the fog, they realized visibility range classification and thusly were able to filter out false targets that don't belong in this range. Even though their experiments were confined within a critically close range (30 m), they paved a way for recovering targets hidden inside fog and smoke, regardless of the attenuation and scattering effects as long as the signal power stays above the designated floor level, because too low a visibility, like below 10 m, blocks the detection almost entirely. Most importantly, the concept of waveform identification brought the Multi-echo technique to the commercial LiDAR markets.

SICK AG company developed an HDDM+ (High Definition Distance Measurement Plus) technology [140], which receives multiple echoes at a very high repetition rate. The uniqueness of the waveform of fog, rain, dust, snow, leaves and fences are all recognizable to their MRS1000 3D LiDAR [180] and the accuracy of object detection and measurement is largely guaranteed. They are also capable of setting a region of interest (ROI), whose boundaries are established based on max & min signal level and max & min detection distance. Such technology provides a very promising solution to the problem of agglomerate fog during heavy rain and other extremely low visibility conditions. Wallace et al. [138] explored the possibility of implementing Full Waveform LiDAR (FWL) in fog conditions. This system records a distribution of returned light energy and thusly can capture more information compared to discrete return LiDAR systems. They evaluated the 3D depth images performance using FWL in a fog chamber at a 41 m distance. This type of LiDAR can be classified as a single-photon LiDAR and 1550 nm wavelength, which Tobin et al. [141] also used to reconstruct the depth profile of moving objects through fog-like high level obscurant at a distance up to 150 m. The high sensitivity and high resolution depth profiling that single-photon LiDAR offers make it appealing in remote, complex and highly scattering scenes. But this raises a question of 1550 nm wavelength and OPA manufacturing difficulties which we will discuss in Sec.11.3.1. Anyway, the idea of waveform identification and multi-echo processing is the main guidance right now in LiDAR perception enhancement and should be able to express more potential if the semiconductor technology can keep its pace with the market demand.

One of the typical LiDAR de-noising works in fog is the CNN-based WeatherNet constructed by Heinzler et al. [181]. Their model trained from both fog chamber data and augmented road data is able to distinguish the clusters in point clouds caused by fog or rain and hence remove them with high accuracy. Lin et al. [143] implemented the nearest neighbor segmentation algorithm and Kalman filter on the point cloud. However, an improvement rate of less than 20% within the 2 m range is considered merely passable. No doubt that the LiDAR and radar combination can tackle fog conditions as well. Qian et al. [139] introduced a Multimodal Vehicle Detection Network (MVDNet) featuring LiDAR and radar. It first extracts features and generates proposals from both sensors, and then the multimodal fusion processes region-wise features to improve detection. They created their own training dataset based on the Oxford Radar Robotcar [75] and the evaluation shows much better performance than LiDAR alone in fog conditions. Mai et al. [142] applied fog to the public KITTI dataset to create a Multifog KITTI dataset for both images and point clouds. They performed evaluation using their Spare LiDAR Stereo Fusion Network (SLS-Fusion) based on LiDAR and camera. By training their network with both clear and foggy data, the performance was improved over a quarter, on the basis of the original performance was reduced by almost a half. In fact, no matter which auxiliary sensor(s) is enlisted, the core of perception enhancement with fusion is to best utilize the data from each sensor and extract useful features and achieve the best result of 1 + 1 >2.

Shamsudin et al. [137] proposed algorithms for fog elimination from 3D point clouds after detection. Clusters are separated using intensity and geometrical distribution and targeted and removed. The restriction is that their environment is an indoor laboratory and the algorithms are designed for building search and rescue robots whose working condition has too low a visibility to be adapted into outdoor driving scenarios where beam divergence and reflectance exist in the far-field. But once again, it reminds people of the de-noising method, namely the de-hazing technique in foggy conditions.

### 5.2.2. Fog in images

Due to the sensitivity of image collecting sensors to external environments, especially under hazy weather, outdoor images will experience serious degradation, such as blurring, low contrast, and color distortion [182]. It is not helpful for feature extraction and has a negative effect on subsequent analysis. Therefore, image de-hazing has drawn extensive attention.

The purpose of image de-hazing is to remove the bad effects from adverse weather, enhance the contrast and saturation of the image and restore the useful features. In a word, estimating the clean image from the hazy input. Currently, existing methods can be divided into two categories. One is non-model enhancement methods based on image processing (Histogram Equalization [148], Negative Correlation [149], Homomorphic Filter [150], Retinex [151], etc.), another is image restoration methods based on atmospheric scattering model (Contrast Restoration [152], Human Interaction [153], Online Geo-model, Polarization Filtering [154]). Although the former can improve the contrast and highlight the texture details, it does not take into account the internal mechanism of the haze image. Therefore, the scene depth information is not effectively exploited and it can cause serious color distortion. The latter infers the corresponding haze-free image from the input according to the physical model of atmospheric scattering. Based on it, a haze model can be described as:

$$I(x) = J(x)t(x) - A(1 - t(x)) \qquad (1)$$

where $I(x)$ is the observed hazy image, $J(x)$ is the scene radiance to be recovered. $A$ and $t(x)$ are the global atmospheric light and the transmission map, respectively. Consider the input $I$ is an RGB color image, at each position, only the three intensity values are already known while $J$, $t$, and $A$ remain unknown. In general, the model itself is an ill-posed [183] problem which means its solution involves many unknown parameters (such as scene depth, atmospheric light, etc.).Therefore, many de-hazing methods will first attempt to compute one or two of these unknown parameters under some physics constraint and then put them together into a restoration model to get the haze-free image.

Until a few years ago, the single image de-hazing algorithm based on physical priors was still the focus. It usually predefines some constraints, prior or assumptions of the model parameters first, and then restores the clean image under the framework of atmospheric scattering model, such as contrast prior [155], airlight hypothesis [156]. However, deducing these physical priors requires professional knowledge and it is not always available when applied to different scenes. With the advance of deep learning theory, more and more researchers introduced this data-driven method into the field.

Chen et al. [145] find that de-hazing models trained on synthetic images usually generalize poorly to real-world hazy images due to the domain gap between synthetic and real data. They proposed a principled synthetic-to-real de-hazing (PSD) framework which includes two steps. First, a chosen de-hazing model backbone is pre-trained with synthetic data. Then, real hazy images are used to fine-tune the backbone in an unsupervised manner. The loss function of the unsupervised training is based on dark channel prior, bright channel prior and contrast limited adaptive histogram equalization.

Considering the problem that the existing deep learning-based de-hazing methods do not make full use of negative information, Wu et al. [147] proposed a novel ACER-Net, which can effectively generate high-quality haze-free images by contrastive regularization (CR) and highly compact autoencoder-like based de-hazing network. It defines a hazy image, whose corresponding restored image is generated by a de-hazing network and its clear image as negative, anchor and positive respectively. CR ensures that the restored image is pulled closer to the clear image and pushed away from the hazy image in the representation space. Zhang et al. [146] employ temporal redundancy from neighborhood hazy frames to perform video de-hazing. Authors collect a real-world video de-hazing dataset containing pairs of real hazy and corresponding haze-free videos. Besides, they propose a confidence-guided and improved deformable network (CG-IDN), in which confidence-guided pre-dehazing module and the cost volume can benefit the deformable alignment module by improving the accuracy of the estimated offsets.

Existing deep de-hazing models have such high computational complexity that makes them unsuitable for ultra-high-definition (UHD) images. Therefore, Zheng et al. [144] propose a multi-guide bilateral learning framework for 4K resolution image de-hazing. The framework consists of three deep CNNs, one for extracting haze-relevant features at a reduced resolution, one for learning multiple full-resolution guidance maps corresponding to the learned bilateral model, and the final one fuses the high-quality feature maps into a de-hazed image.

Recently, in image de-hazing, an unpaired image-to-image translation that aims to map images from one domain to another come into focus. It gets boosted by generative adversarial networks (GAN) that have the ability to generate photorealistic images. CycleGAN [184], DiscoGAN [185], and DualGAN [186] are three pioneering methods, which introduce the cycle-consistency constraint to build the connection. Note that, this method does not require a one-to-one correspondence between source and target, which is more suitable for de-hazing. Because it is almost impossible to collect different weather conditions while keeping the background unchanged at the pixel level, considering that the atmospheric light changes all the time. Engin et al. [187] proposed Cycle-Dehaze which is an improved version of CycleGAN that combines cycle consistency and perceptual losses in order to improve the quality of textural information. Shao et al. [188] proposed a domain adaptation paradigm that introduces an image translation module that translates haze images between the real and synthesis domain. Such methods are just getting started, and the results of de-hazing are often unsatisfactory (artifacts exist). But its feature does not require paired images to have the potential to build more robust models.

Although the field has approached maturity, the mainstream methods still use synthesis data to train models. Because collecting pairs of hazy and haze-free ground-truth images need to capture both images with identical scene radiance, which is almost impossible in real road scenes. Inevitably, the existing de-hazing quality metrics are restricted to non-reference image quality metrics (NRIQA) [189]. Recent works start to collect haze datasets utilizing a professional haze/fog generator that imitates the real conditions of haze scenes [190], or multiple weather stacking architecture [191] which generates images with diverse weather conditions by adding, swapping out and combining components. Hopefully, this new trend could lead to more effective metrics and boost the existing algorithms to deploy on the ADS.

### 5.2.3. Gan-based de-hazing model experimental evaluation

We did an evaluation on our own gan-based model as shown in Fig.13. Specifically, on the architecture of CycleGAN [184], we added weather layer loss and spatial feature transform technique to disentangle hazy images from the front hazy layer, which keeps the background content in the de-hazing process to a maximum extent. The model is trained on Cityscapes and Foggy Cityscapes datasets [193]. After training the GAN based de-hazing model, we first apply it to the hazy input. Then we use the state-of-the-art pedestrian detector in CityScapes dataset to verify the significance of de-hazing. The results show that the amount of valid detection is increased after haze removal, especially the ones that are partially obscured in the back. For more details about this GAN-based de-hazing model, please refer to [192].

### 5.3. Snow
#### 5.3.1. Snow covering

One branch of perception in snow is pathfinding due to the snow covering, which already caught the attention of autonomous robots at an early stage. Yinka et al. [157] proposed a drivable path detection system in 2014, aiming at extracting and removing the rain or snow in the visual input. They distinguish the drivable and non-drivable paths by their different RGB elements values since the white color of snow is conspicuous compared to road surfaces, and then apply a filtering algorithm based on modeling the intensity value pixel of the image captured on a rainy or snowy day to achieve removal. Their output is in mono color condition and the evaluation based on 100 frames of road pictures shows close to 100% in pathfinding. Although the scenario is rather simple where only some snow is accumulated by the roadsides, this lays a good foundation for ADS when dealing with the same problem in snow conditions.

Vachmanus et al. [159] extended this idea into the autonomous driving semantic segmentation task by adding thermal cameras into the modality. RGB camera input might not be enough to represent every pertinent object with various colors in the surroundings, or pedestrians involved in the snow driving scenario, which happens to be the thermal camera's strong point. Their architecture contains two branches of encoders, one for RGB camera and thermal camera each to extract features from their own input. The temperature feature in the thermal map perfectly supports the loss of image element due to the snow and the fusion model successfully improves snow segmentation compared to not only RGB camera alone, but several other state-of-art networks, based on the validation on several datasets including Synthia and Cityscapes. This network is very suitable for automated snowplows on roads with sidewalks, which serves beyond the traditional autonomous driving purpose and could be of real commercial use earlier than AV.

Furthermore, Rawashdeh et al. [158] include cameras, LiDAR and radar in their CNN sensor fusion for drivable path detection. This multi-stream encoder-decoder almost complements the asymmetrical degradation of sensor inputs at the largest level. The depth and the number of blocks of each sensor in the architecture are decided by their input data density, of which camera has the most, LiDAR the second and radar the last, and the outputs of the fully connected network are reshaped into a 2-D array which will be fed to the decoder. Their model can successfully ignore the lines and edges that appeared on the road which could lead to false interpretation and delineate the general drivable area. Such fusion modality certainly can do more than countering snow conditions but other low-visibility scenarios and people are working towards more channels in the network since LiDAR and radar can provide more types of information like reflected beam numbers and velocities.

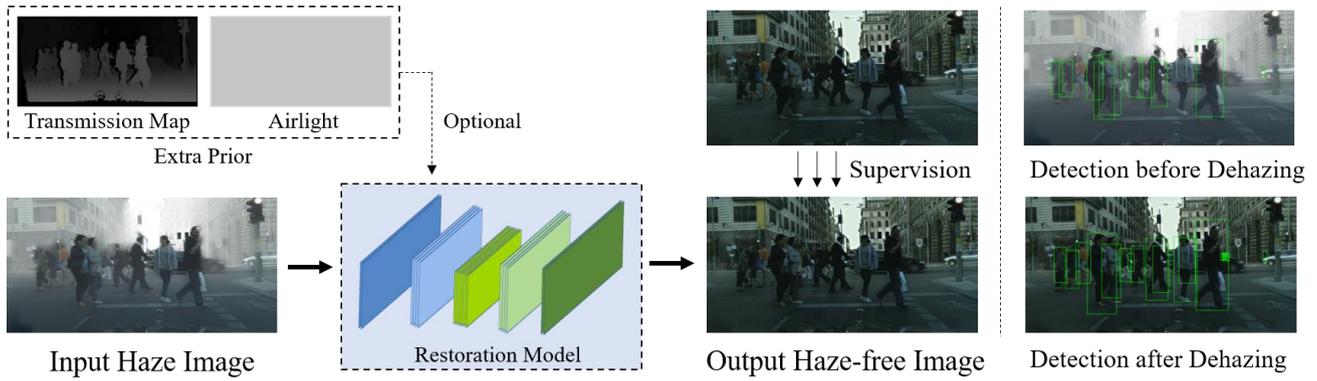

**Figure 13:** Illustration of de-hazing methods based on atmospheric scattering model [192].

*5.3.2. Snowfall*

The other branch of perception in snow is the degradation of signal or image clarity caused by snowfall, just like raindrops, only falling at a slower speed but with larger volume. As for snowfall, the coping method once again returns to the de-noising technique, even for LiDAR point clouds. Charron et al. [160] extensively explained the deficiency of 2D median filter and conventional radius outlier removal (ROR) before proposing their own dynamic radius outlier removal (DROR) filter. As snowfall is a dynamic process, only the data from the lasers pointing to the ground is suitable for a 2D median filter while it's not necessary from the beginning. The data are quite sparse in the vertical field of view above ground and the 2D filter couldn't handle the noise point removal and edge smoothing properties well. Hence 3D point cloud ROR filter is called for. This filter iterates through each point in the point cloud and examines the contiguous points within a certain vicinity (search radius), and if the number of points found is less than the specified minimum ($k_{min}$), then this point would be considered as noise and removed, which fits the pattern of snowfall where snowflakes are small individual solid objects. The problem is directly implementing this filter in the three-dimensional sense would cause the undesirable removal of points in the environment far away and compromise the LiDAR's perception ability in terms of precognition, as shown in Fig.14(d). To prevent this problem, Charron's group applied the filter dynamically by setting the search radius of each point ($SR_p$) according to their original geometric properties, as shown in Eq.2, and successfully preserved the essential points in the clouds far away from the center (6m - 18m) while removing the salt and pepper near the center (within 6m) in the point clouds with a precision improvement of nearly 4 times of normal ROR filters, as shown in Fig.14(e).

$$SR_p = \beta_d (r_p \alpha) \qquad (2)$$

$r_p$ is the range from the sensor to the point $p$, $\alpha$ is the horizontal angular resolution of the LiDAR, and the product of $(r_p \alpha)$ represents point spacing, which is expected to be computed assuming that the laser beam is reflecting off a perpendicular surface. So the multiplication factor $\beta_d$ is meant to account for the increase in point spacing for surfaces that are not perpendicular to the LiDAR beams [160].

On the other hand, Park et al. [163] proposed a low-intensity outlier removal (LIOR) filter based on the intensity difference between snow particles and real objects. It can also preserve important environmental features as the DROR filter does, but somehow maintain more points in the cloud than DROR because LIOR's threshold is more targeted based on the subject's optical properties. It could be an advantage in accuracy given the right circumstances.

The de-snowing technique for camera works in a similar way with de-hazing. Zhang et al. [162] proposed a deep dense multi-scale network (DDMSNet) for snow removal. Snow is first processed by a coarse removal network with three modules, pre-processing module, a core module and a post-processing module, each containing a different combination of dense block and convolutional layers. The output is a coarse result where the negative effect of falling snow is preliminarily eliminated and is fed to another network to acquire semantic and geometric labels. The DDMSNet learns from the semantic and geometry priors via self-attention and generates clean images without snow. The interesting part is that they use Photoshop to create large-scale snowy images out of Cityscapes and KITTI datasets to do the evaluation. Despite the fact that this is indeed totally capable of performing state-of-art snow removal, it's still necessary to introduce advanced methods of simulating photo-realistic snow images.

Von Bernuth et al. [161] generate snow in three steps: first, reconstruct the 3D real-world scene with depth information in OpenGL; then snowflakes are distributed into the scene following physical and meteorological principles, including the motion blur that comes from wind, gravitation or the speed of vehicle displacement; finally, OpenGL renders the snowflakes in the realistic images. The depth information is critical for reconstructing the scene, so the images are either gathered from stereo cameras or other sensors in the real world like the two datasets mentioned above, or from simulators like Vires VTD or CARLA whose depth information is perfectly quantifiable. The snowflakes have two forms: flat crystal as if in 2D, and thick aggregated flakes constructed by three pairwise perpendicular quads in 3D, which ensure the synthetic snow looks like reality as close as possible. Comparison of such methods of snow generating shows a stunning resemblance with real-world snowy images. No doubt that de-noising with synthetic snowy and foggy images can help the machine learning process and benefit camera perception enhancement in adverse weather conditions to a great extent.

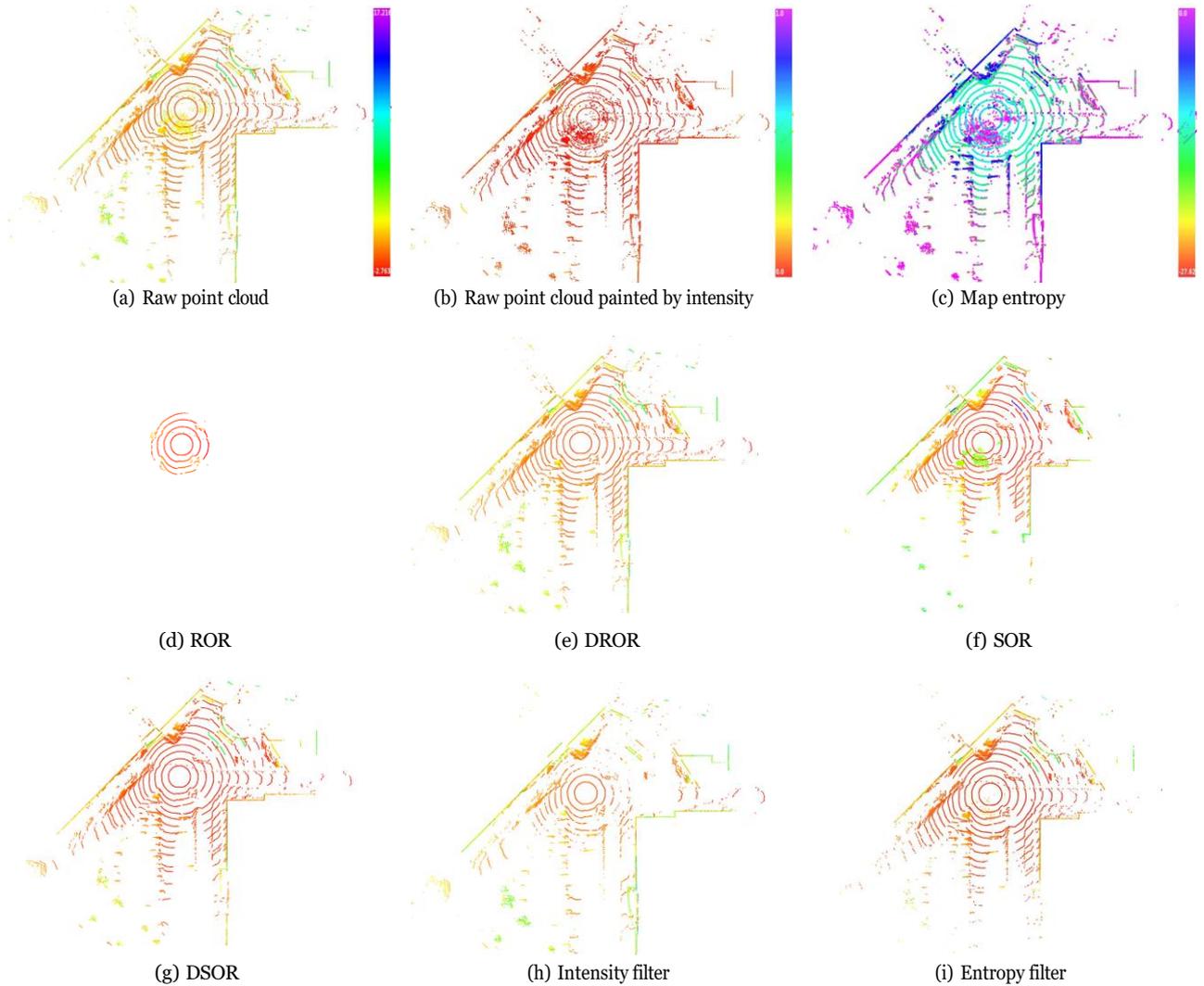

**Figure 14**: Velodyne HDL-32 LiDAR point clouds in snowfall conditions with different de-noising methods, reproduced using Canadian adverse driving conditions dataset (CADCD) [58]. (a) Raw point cloud painted by height; (b) Raw point cloud painted by intensity; (c) Map entropy painted by entropy; (d) to (i) are painted by height (Z aixs), and share the same color scale as (a). Point clouds denotations: buildings (peripheral regular shapes); ground points (rings); curb points and snow points (scattered points around the center).

### 5.3.3. *Point cloud de-snowing validations*

In this section, we present experimental validations on some common point cloud de-snowing filters including the ROR and DROR from above and an entropy filter we proposed by ourselves. In addition to Fig.14(d) and 14(e) we produced for illustration, we also present the raw point cloud in intensity scale and an intensity filter; the map entropy of the raw point cloud and an entropy filter; as well as testing and validating SOR and DSOR filters in Fig.14, as comparisons for better references. All the validations are based on a scene captured from the Canadian adverse driving conditions dataset (CADCD) [58] where buildings, trees, and parked cars are widely present, as shown in Fig.14(a) and 14(b).

Fig.14(f) shows the Statistical Outlier Removal Filter (SOR), which calculates the mean and standard deviation of the distance to its $k$ nearest neighbors when iterating each point. The threshold $T$ for filtering is computed as:

$$T = \mu + \sigma\beta_s \qquad (3)$$

where $\mu$ is the global mean of the distances from all points to their $k$ nearest neighbors; $\sigma$ is the global standard deviation of the distances; and $\beta_s$ is a specified multiplier parameter. It can be seen from Fig.14(f) that ROR's flaw has been largely improved but at the cost of de-noising performance.

With Dynamic Statistical Outlier Removal (DSOR), as shown in Fig.14(g), the advantages of DROR and SOR are combined, i.e. the filter threshold of SOR is dynamically changed with range [194]. The dynamic threshold $T_d$ is set by:

$$T_d = r(T r_p) \qquad (4)$$

where $T$ is from Eq.3 and $r_p$ is the distance of every point from the sensor, same as in Eq. 2. $r$ is a multiplicative factor for point spacing. Larger $r$ leads to a milder filter. It turns out the performance is no less than DROR in both de-noising and preserving environmental features, and even faster in terms of computing.

Fig.14(h) is a direct intensity filter where all the points with intensity values outside of the interval of [0.03, 0.15] are filtered out (intensity varies from [0,1]). It can be seen that the result is somehow acceptable but the problem is how to acquire the exact interval in each scene that can both filter out snowfall and keep objects. Therefore, the practical use of the intensity filter is limited.

Fig.14(c) shows the entropy representation of the original raw point cloud scene. The entropy $h$ of a certain point $q_k$ in the point cloud is computed by:

$$h(q_k) = \frac{1}{2}ln\,|2\pi e\Sigma(q_k)| \qquad (5)$$

in which $\Sigma(q_k)$ is the sample covariance of mapped points in a local radius $r$ ($r = 0.25$ m in our case) around $q_k$. Fig.14(i) is the result after the solitary points (points with less than 15 neighbors) being punished [195]. As we can see, when the snowfall is very dense, the entropy filter is still having trouble filtering them all.

## 5.4. Light related
### 5.4.1. Strong light and glare

Notwithstanding the severe influences of strong light and glare on AV, there is very limited literature specifically targeting the solution to light-related problems. Back in 2014, Maddern et al. studied the effect to an AV caused by light condition changes during the 24 hours of a day, and managed to improve the performance and robustness of vision-based autonomous driving by implementing illumination invariant transform, which removed almost all variation due to sunlight intensity, direction, spectrum and shadow present in the raw RGB images [172]. Commonly the idea is to count on the redundancies and robustness of certain fusion modalities that are equipped with sensors agnostic to strong light, which leaves glare detection or to say the awareness of strong light, as the job.

Yahiaoui et al. [164] developed their own sunshine glare dataset in autonomous driving called Woodscape, including situations like direct sunlight in the sky or sun glares on dry roads, road marks being wiped off by sun glares on wet roads, sun glares on reflective surfaces, etc. The glare is detected by an image processing algorithm with several processing blocks including color conversion, adaptive thresholding, geometric filters, and blob detection, and trained with CNN network.

### 5.4.2. Reflections and shadows

Glare and strong light might not be removed easily, but reflections in similar conditions are relatively removable with the help of the absorption effect [165], reflection-free flash-only cues [167], and photo exposure correction [169] techniques in the computer vision area. The principle follows reflection alignment and transmission recovery and it could relieve the ambiguity of the images well especially in panoramic images which are commonly used in ADS [168]. It's limited to recognizable reflections and fails in extremely strong lights where image content knowledge is not available. A special reflection is the mirage effect on hot roads. It has a weakness that the high-temperature area on the road is fixed and that fits the feature of a horizon [171]. Kumar et al. [170] implemented horizon detection and depth estimation methods and managed to mark out a mirage in a video. The lack of mirage effects in dataset makes it hard to validate the real accuracy.

The same principle applies to shadow conditions as well, where the original image element is intact with a little low brightness in certain regions [166]. Such image processing uses similar computer vision techniques as in previous paragraphs and can also take the route of first generating shadows then removing them [173]. The Retinex algorithm can also be used for image enhancement in low-light conditions [174]. Nevertheless, reflection and shadow conditions do not threaten a mature ADS because of the presence of LiDAR and radar.

### 5.4.3. Thermal imaging validations

In order to show the superiority of thermal imaging over normal RGB camera, especially in adverse light conditions such as the strong light condition, we did some validations of a couple of thermal imaging enhancement algorithms based on the same scene in Fig.6.

We can see from Fig.15 that under direct strong light the highlighted objects were detected at 40 m from the Xenon light source when illuminance is maximum (200 klx) but in all cases there are no enough measurement points to achieve recognition, but thermal camera still remains part of the abilities to distinguish the rectangle board beneath the light source which normal camera couldn't. In addition, we applied multiscale retinex transformation [196], and the parvo cellular representation of a bio-inspired retina method [197] to the thermal image to enhance perception. The multiscale retinex transformation has the ability of image color restoration and contrast enhancement, while the parvo cellular retina model provides accurate structuring of video data by noise and illumination variant removal and static & dynamic contour enhancement.

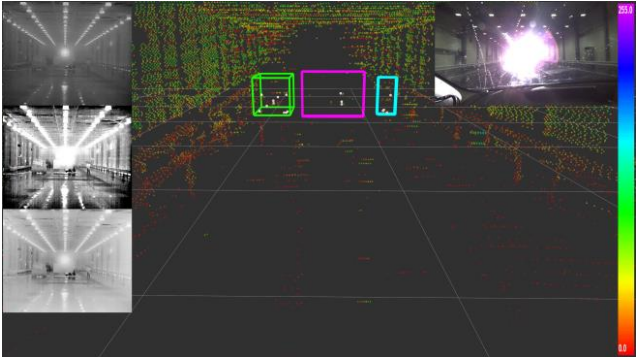

**Figure 15:** Strong light affects negatively the LiDAR (point cloud colored by intensity) and RGB camera (top right inset). In the experiment, the vehicle is located 40 m from the Xenon light source with a peak illuminance of 200 klx. Objects are barely detected: four 3D points for the mannequin (cyan box), three points for the reflective targets (magenta box), and ten points for the black vehicle (green box), classification is not possible. However, the thermal camera is resilient to illumination and the objects are clearly discernible (top left inset). Multiscale retinex transformation (left middle inset). Parvo cellular representation of a bio-inspired retina method (left bottom inset).

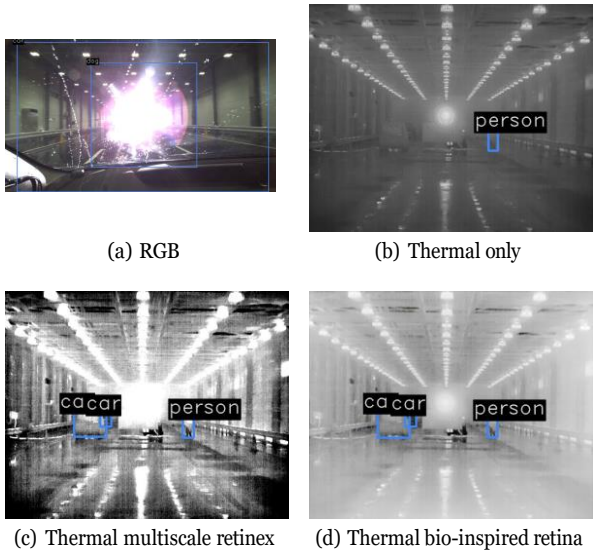

**Figure 16:** Object classification results using Yolo3 on RGB and thermal camera images with strong light. (a) RGB camera; (b) thermal; (c) multiscale retinex enhancement; (d) bio-inspired retina enhancement.

Furthermore, we tested and compared the Yolo3 [198] object classification results in Fig.16 among RGB camera and each one of the thermal imaging from above. Three classes are defined: person (the mannequin), car 1 (the vehicle in front), and car 2 (the vehicle at the back). 10 seconds of camera frames were analyzed while the car approached the strong light source. It can be seen from Fig.16(a), a normal visible camera is almost blinded and the light source halo is recognized as "dog". The recall rates are all zero for the three classes. In Fig. 16(b), partial ability is regained where the person can be recognized but not the cars from behind. The recall rates are only 44.3%, 0% and 60% respectively for the three classes. Multiscale retinex enhancement and bio-inspired retinex enhancement successfully captured all three elements with good accuracy. Multiscale retinex [196] enhancement (Fig.16(c)) has recall rates of 58.6%, 30% and 62.9% respectively; and bio-inspired retina [197] enhancement (Fig.16(d)) has 67.1%, 45% and 78.6% respectively.

### 5.5. Contamination

As we summarized in Table 1, contamination influences the perception of ADS sensors in a fierce way, like an invasion of the sensor's line of sight. For example, the contamination effect on the backup camera is shown in Fig.17. As a result, the robustness and adaptability of the system are facing a rigorous test. Uřičář et al. [175] from Valeo created a dataset called SoilingNet having both opaque and transparent soiling [176], and developed a GAN-based data augmentation for camera lens soiling detection in autonomous driving. Different from rain or snow, the general soiling is normally considered as opaque or semi-transparent, so a complementary sensing method might not be able to perform with enough accuracy. Once again, the Valeo group starts from the artificial soiling image generations as the impossibility of acquiring of both the soiled image and the same clean image in real driving conditions. The CycleGAN network would generate an image with a random soiling pattern, which provides a blurred mask obtained from the semantic segmentation network applied with a Gaussian smoothing filter on the generated soiled image, and finally, the synthetic version of the soiled image is composed with the original image and the soiled pattern estimated via the mask. The degree of similarity is very close to the real soiled effect shown in Fig.17(c). The only problem is that CycleGAN does not have the restraint on soiling a designed region of the image but transforms the whole image, so they apply restrictions on the mask area only and modify the network to a new DirtyGAN. Furthermore, they used this DirtyGAN to generate a Dirty dataset based on their previous dataset Woodscape as mentioned in the last paragraph and the degradation evaluation based on the Cityscapes dataset is proven well. Although the removal or interpretation of the soiled image was not discussed in this work, it sure provides a possibility of the same training approach as de-hazing and de-noising. To say the least, it's now realistic to call upon mechanical devices like wipers or sprayers once ADS is able to detect contamination.

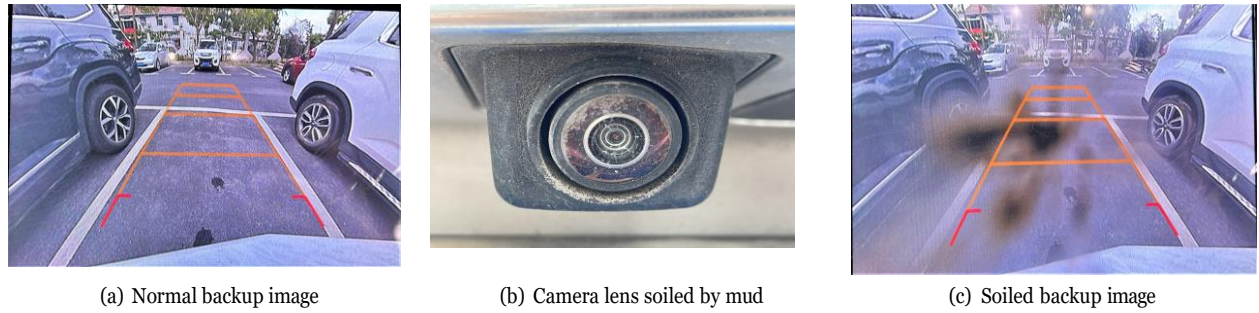

(a) Normal backup image    (b) Camera lens soiled by mud    (c) Soiled backup image

**Figure 17**: Contamination effect on a Cadillac XT5 back-up camera. The mud contamination is formed naturally from off-road driving after rain. Vehicle testing and images courtesy of Mr. Dawei Wang, Pan Asia Technical Automotive Center Co., Ltd.

Trierweiler et al. [84] made an automatic control over the wiper and nozzle based on the same line of thought, only their detection is based on the total internal reflection on the windshield. Due to the difference in reflectivity and intensity distribution between liquid, like raindrops, and solid, like dust, they are able to differentiate the physical status of the pollution on the cover glass of the ADS sensors by splitting the out-coupled intensity from the red light diffuser into two sections. Compared to current rain sensors who can only trigger wipers, this model knows whether it's necessary to trigger the nozzle to clean up the pollution. One concern is that the windshield of a car might not maintain a uniform curvature everywhere, so the practicability of the installation of such a detection device and its operation consistency remains challenged. Though it is still unclear if LiDAR or camera can interpret the affected data caused by contamination on themselves, one conventional plain solution is better than none.

SICK sensor company's microScan 3 safety laser scanner has a voice in this matter as it can resist severe contamination [199]. This close proximity safety monitor scans a $275°$ angle with 845 nm lasers and can work fine in outdoor weather conditions using the HDDM+ technology as mentioned before [140]. The highlight of this sensor is that even with contamination like sawdust adhering to the sensor emitter cover window, it can still detect the presence of objects in close proximity without false alarm as a benefit of the incredibly wide scanning angle and the ability to emit multiple pulses with a $0.1°$ angular resolution. Such a sensor possesses the potential of serving as the final barrier between AV and obstacles when contamination has conquered almost every sensor in extremely adverse conditions.

## 6. Classification and Assessment

Along the pipeline of how ADS work, after the sensors have done collecting information, it's time for the "brain" to make a judgment on the driving condition and its current risk level. A summary of methods for classification including weather classification, visibility classification and risk assessment is listed in Table 4.

### 6.1. Weather classification

Perception enhancement fundamentally enables ADS to navigate through various inclement weather conditions, but it mainly focuses on how to ignore the presence of weather or compensate for the negative effects. Weather classification works from the opposite angle but is not unserviceable. It's equally important to be aware of the current weather condition as being able to see through, and only with proper weather classification abilities can ADS make targeted decisions with high confidence and accuracy. Here we focus on the local-scope weather classification for individual AVs, because macroscopic weather classification, like weather radar, does not have the density and resolution to reflect actual weather effects that are being applied to one car.

At first, weather classification dwelt at few-class weather classification like distinguishing clear or not [200] on single images. Further machine learning techniques like kernel learning achieved multi-class weather classification like sunny, rain, fog and static snow. At this stage, the classification task is realized by setting classifiers with the unique features of each kind of weather. Sunny features come from the clear sky region of a picture and form a highly multi-dimensional feature vector; when sky elements are not included in the picture, a strong shadow region with confident boundaries becomes the indicator of sunny condition. Rain streak is hard to capture in images so HOG features are extracted from the image to be the rain feature vector. Falling snow is considered as noise and pixels with certain gray levels are defined as snowflakes. Haze is determined by dark channels, where some pixels have very low intensities in at least one color channel which is the dark channel [201]. With the development in AI technologies, machine learning neural networks such as deep CNN are used by Elhoseiny et al. [202] in this task to enhance feature extraction and learning performance. Al-Haija et al. [203] came up with a powerful ResNet-18 CNN network including a transfer learning technique to do multi-class weather classification based on the pre-training of multi-class weather recognition on ImageNet dataset. However, the class set in this network is still restricted to sunrise, shine, rain and cloudy, whose impacts on ADS are not obvious or representative. Dhananjaya et al. [204] tested the ResNet-18 network on their own weather and light level (bright, moderate and low) dataset and achieve a rather low accuracy, which means traditional image weather classification is still not saturated. In order to serve weather classification for autonomous driving purposes, fine sorted and precise classification is needed, with the possibility of going beyond camera images only.

Heinzler et al. [205] achieved a pretty fine weather classification with multi-echo LiDAR sensor only. The point cloud is firstly transformed into a grid matrix and the presence of rain or fog can be easily noticed by the occurrence of secondary echos on objects. Then, different from recording the echoes of each kind of condition, the mean distance of each echo and their mathematical properties like variance are used for detailed classification as the covariance matrices are influenced by different levels of rain or fog and the change in the point cloud or to say the matrix is visible. Nearest Neighbor classifier (kNN) and a Support Vector Machine (SVM) are applied as classifiers and rain classification is largely improved. It can be imagined that the test result might not be as good when using a LiDAR sensor with a smaller vertical FOV due to the insufficient number of points and also in dynamic scenarios compared to static scenes. That means this method still has its reliance on controlled environments and the robustness might not meet Level 4 or higher autonomy requirements.

Dannheim et al. [206] did propose to use the fusion data from both LiDAR and camera to do weather classification several years before. Their main classifier was based on the intensity difference generated by the backscattering effect of rain and fog and no neural network was mentioned in their image processing, which doesn't seem effective in today's point of view. So it might be a good idea to combine both advanced image detection and LiDAR data processing mentioned above to realize weather classification and guarantee accuracy and robustness in adverse conditions. If Level 3 autonomy were still on the table of the current market, a human-machine interface that prompts confirmation of basic weather classification results to the vehicle operator might be the most reliable way for the time being, just like when the navigation systems ask approval of route changes. Since less human intervention is the direction in which we are working, the weather classification module is necessary for a mature ADS.

Weather classification done by sensors equipped on V2X or V2I facilities works the same way, but with an extra step of data transmission. Details of V2X will be talked about in Sec. 11.2.3.

### 6.2. Visibility classification

There is another angle of classification in adverse conditions: visibility. This definition was first invented as the subjective estimation of human observers. In order to measure the meteorological quantity, or to say the transparency of the atmosphere, the meteorological optical range (MOR) [207] is defined objectively. In the context of autonomous driving, when visibility is quantified as specific numbers, it normally means MOR. For example, each distinct version of fog scenarios in the Foggy Cityscapes [193] dataset is characterized by a constant MOR. As weather conditions often bring visibility degradation of different levels, it's helpful to gain awareness of visibility dropping for ADS to avoid detection errors and collisions in advance. Traditional visibility classification was usually conducted in facilities like airports where human estimation and commercial instruments take the responsibility [208]. Manual observation of the presence of weather like fog and the range of visibility is considered the most accurate and reliable. Optical visiometers like transmissometers and scatterometers are also commonly used for precise visibility measurement in a rather long range. However, these two approaches are not fit to be applied to the autonomous driving area, so sensors like LiDAR and camera have to take over. It is possible to estimate the visibility range in foggy conditions by profiling the LiDAR signal backscattering effect caused by the tiny droplets, but as mentioned in the previous context, it requires extremely fine-tuned LiDAR power to adapt to the fickle variables, so its practical application is limited. Currently, visibility classification largely relies on camera-based methods with neural networks [209] and is divided by range classes with intervals of dozens of meters while seldom gives exact pixel-wise visibility values [210]. Considering the low cost and the irreplaceable status of cameras in ADS, it is also well researched.

Chaabani et al. [209] initially used a neural network with only three layers: feature vector image descriptor as input, a set of fully interconnected computational nodes as a hidden layer, and a vector corresponding to the visibility range classes as output. They used the FROSI (Foggy ROad Sign Images) synthetic dataset [211] [212] for evaluation and were able to classify the visibility from below 60 m to larger than 250 m with a spacing of 50 m. They later improved such a network with the combination of deep learning CNN for feature extraction and an SVM classifier [213]. The new network used the AlexNet architecture [214] which is consisted of five convolution layers, three maxpool layers and three fully connected layers. The overall recall, precision and accuracy all reached the state-of-art level and can be used on not only car on-board cameras but roadside cameras which shows further potential in future IoT systems.

Duddu et al. [215] proposed a novel fog visibility range estimation algorithm for autonomous driving applications based on a hybrid neural network. Their input consists of Shannon entropy [216] and image-based features. Each image captured by the 50-degree-FOV camera is divided into 32 by 32 pixel blocks and the Shannon entropy of each block is calculated and then mapped to corresponding image features extracted from a series of convolutional layers along with maxpool layers, which output three visibility classes: 0 - 50 meters, 50 - 150 meters, and above 150 meters. They created their own fog dataset with BOSCH range finder equipment as ground truth to establish the network architecture and the synthetic dataset FORSI is used for public benchmarking. The overall accuracy of 85% and higher enables ADS to set up thresholds of dangerous low visibility.

**Table 4**
Summary of methods on classification, localization & mapping, and planning & control with regard to weather

| Classification | | | Localization and Mapping | | | | Planning and Control | |
|---|---|---|---|---|---|---|---|---|
| Weather classification | Visibility classification | Risk assessment | SLAM | A-priori map | LiDAR | Other | Planning | Control |
| [200] [201] [202] [203] [204] [205] [206] | [209] [211] [212] [213] [214] [215] [217] [219] [218] [220] | [221] [222] [223] | [224] [225] [226] [227] | [228] [229] [230] [231] [44] | [232] [233] [122] [234] | [75] [235] [236] | [237] [238] [239] | [240] [241] [242] [243] [244] |

There are also other similar models like the feed-forward back-propagation neural network (BPNN) [217] using data collected from weather monitoring stations as input, that can predict the visibility ranges with much smaller spacing at a road-link level. It is unclear whether mobile weather stations equipped on cars are capable of completing visibility classification in real time, but sophisticated sensor fusion could be necessary for conditions beyond fog like snow and rain. Considering a visibility classification module that can provide warning messages to human drivers is useful in current ADAS, it's possible that this technique goes to implementation earlier than full autonomy.

As a matter of fact, there is a correlation between weather and visibility in climatology and research has been done since decades ago about the correspondence between how far a driver can see and precipitation rates [70]. If a detailed correspondence is available, then probably we only need to detect the rain or snow rate rather than classifying visibility through computer vision technique. However, this type of one-to-one correspondence chart works at a rather longer range because most of the precipitation would not bring the road visibility down to a level below 1 km, far away from the current AV's visual concern. The visibility crisis mostly comes from the water screen and mist during rain and it sometimes doesn't depend on the rain rate only. It seems like that only thick fog or pollution smog or sandstorm whose visibility is normally below a couple of hundred meters is meaningful to be considered in this direct way. Sallis et al. [218] thought of using vehicular LiDAR's backscatter effect to detect air pollution and fog intensity, but visibility are not considered at that time. Miclea et al. [219] came up with a creative way by setting up a 3-meter-long model chamber with a "toy" road and model cars in it which can be easily filled with almost-homogeneous fog. They successfully identified the correlations between the decrease in optical power and the decrease in visual acuity in a scaling fog condition. Furthermore, Yang et al. [220] managed to provide a promising prediction of a 903 nm NIR LiDAR's minimum visibility in a fog chamber by determining whether the detecting range of an object with a known distance is true or noisy. Even though these are still indirect methods, perhaps with future IoT or V2X technology, real-time visibility classification could be realized.

### 6.3. Risk assessment

When it comes to driving, visibility is merely a direct perception. What really determines the way we drive including speed control and course changing is the risk level assessed by our brain based on the perception inputs. There is an important definition for both human driving and AV called Stopping Sight Distance (SSD). According to the American Association of State Highway and Transportation Officials (AASHTO), SSD is the sum of reaction distance (assume 2.5 s reaction time for humans) and braking distance (deceleration rate $3.4\ m/s^2$) [245]. For reference, the designed SSD is normally 85 m at the speed of 60 km/h. Considering machines should respond quicker than humans, a 0.5 s reaction time (about at least 15 camera frames, or 5 LiDAR frames, or 20 radar frames) for AV is assumed in literature [221], which means a much shorter SSD for AV and a much larger challenge on safety. Shalev-Shwartz et al. [222] from Mobileye proposed the Responsibility-Sensitive Safety (RSS) model for AV safety. The top 1 rule is "Do not hit the car in front" with longitudinal distance control.

All the rules mentioned above stand for a risk assessment based on quantitative visibility estimation under normal or ideal driving conditions. When inclement weather strikes, two possible scenarios may pose threats to these basic safety rules: (1) impaired sensors like soiled cameras lose partial visibility or direct sightlines to some of the surroundings, (2) severe weather conditions like water mist or dense fog degrade the visibility to a critical level which is lower than designed SSD. Such scenarios largely increase the uncertainty of the input variables and discredit the preset safety standards and thresholds. Similar to complex road scenes when pedestrians or cyclists are hard to be detected or predicted by AV, having enough dense information like LiDAR point clouds at questionable locations to investigate the risk level based on visibility in advance is meaningful [223]. Rarely any literature focuses on the risk assessment specifically in the autonomous driving area, but a risk assessment or to say reliability analysis module should be incorporated in all the decision-making and control processes related to visibility to ensure the utmost safety for AV in any possible and unpredictable conditions.

## 7. Localization and Mapping

The awareness of an ego vehicle's own location is as important as knowing other elements' locations in the surrounding environment. The most common methods currently involved in localization are the Global Positioning System and Inertial Measurement Unit (GPS-IMU), SLAM, and state-of-the-art a-priori map-based localization. Normally the one with a higher accuracy comes with a higher cost, but we are going to focus on their robustness in weather conditions alone. A summary of localization methods in adverse weather conditions is shown in Table 4.

Achieving high-accuracy localization with merely GPS or INS (internal navigation system) could be a challenge because they generally have a double-standard deviation of over 1 meter on most roads [246] despite the assistance of wheel odometry, let alone harsh and slippery road conditions. Still, Onyekpe et al. [236] exploited deep learning and proposed the Wheel Odometry neural Network (WhONet) to evaluate the uncertainties in the wheels displacement measurement and find the compensating factor. The tests against wet and muddy roads in the Coventry University Public road dataset for Automated Cars (CUPAC) showed up to over 90% reduction in positioning error compared to physical models, in GNSS-deprived environments. This network is not specifically targeting weather scenarios but mostly on changes to the tires, so the actual performance in weather-induced road conditions is yet to be verified.

### 7.1. Simultaneous localization and mapping

The same-time online map making and localization method is widely deployed in robotics and indoor environments, so it doesn't face the challenge of wet weather very often in the outdoor. However, the change of feature descriptors across seasons compromises SLAM's accuracy to some extent. Besides season changing, weather induced effects including tree foliage falling and growing and snow-covered ground are also part of the reasons. To address the robustness problem of SLAM, Milford et al. [224] proposed to recognize coherent navigation sequences instead of matching one single image and brought the SeqSLAM as one of the early improvements of SLAM in light, weather and seasonal changes conditions. SeqSLAM has a weakness of assuming well alignment in different runs which could result in poor performance with uncropped images or different frame rates. Naseer et al. [225] took it to a further step by first using deep convolutional neural network (DCNN) to extract global image feature descriptors from both given sequences, then leveraging sequential information over a similarity matrix, and finally computing matching hypotheses between sequences to realize the detection of loop closure in datasets from different seasons.

Wenzel et al. [226] collected in several European cities under a variety of weather and illumination conditions and presented a Cross-Season Dataset for Multi-Weather SLAM in Autonomous Driving called 4Seasons. They showed centimeter-level accuracy in reference poses and also highly accurate cross-sequence correspondences, on the condition of good GNSS receptions though.

As robust as visual SLAM may have become, the map's insufficiency in necessary information agnostic to appearance changes is still one of the major problems of SLAM. The localization drift over time and the lack of viability of map in every driving condition also hinder SLAM from navigating for long distances, which makes it less competitive compared to pre-built map based localization in autonomous driving [227].

### 7.2. A-priori map

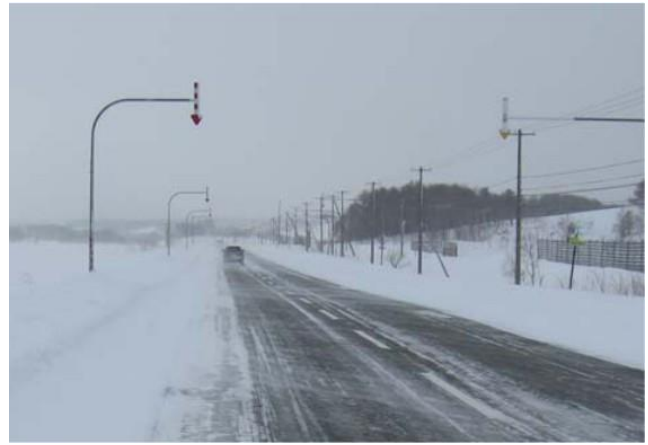

**Figure 18**: Down-pointing arrow marking the edges of the road covered by snow in Hokkaido, Japan. Image courtesy of Dr. Atsushi Nishimura of Snow and Ice Research Team of CERI (Civil Engineering Research Institute for Cold Region)[247].

The reference pre-built map that is used for comparing and matching the online readings for localization purposes is called the a-priori map, which is currently the main force in ADS localization. A key method in the assurance of the accuracy of a-priori map-based point cloud matching is landmark searching. The most straightforward way to teach AV how to drive right now is by letting AVs think and behave like humans do. A typical example of how humans deal with low certainty localization is the snow pole "Yabane" (down-pointing arrow on a pole) in Hokkaido, Japan, as shown in Fig.18. The arrows point to the edge of the road and are either flashing or light reflective at night, so this gives the driver a rough idea about their relative position on the road and keeps them from wandering and away from the curb. This concept is important because the occlusion caused by accumulated snow has much more impact on localization than snow precipitation [248].

At the early stage of self-driving, RFID (radio frequency identification) technology was used for robots localization in controlled environments and ODDs [228]. Geographic information embedded in the RFID tags by the route checkpoints can be read by the antennas when the robot reaches within the one or two meters range. Such methods are still being used by AGVs (Automatic Guided Vehicles) in factories or warehouses [229]. With V2I communications available, AVs can enjoy the same localization privilege [230]. Given that the markers' presence is considered fixed, the basic idea of landmark searching in AV localization emerges as time requires. Enterprises like Ford work closely with companies major in detailed 3D a-priori maps, and their automated Fusion sedan would seek for markers like stop signs or other signposts by the roadsides with either camera or LiDAR, and sense its ego-position by calculating its relative distance against those markers in the a-priori map [231]. As simple as it may sound, this still gives Ford the ability to locate their own AV as accurately as in centimeters with no extra hardware cost. However, the potential cost and difficulty lie in the constant updating and maintenance of the a-priori map and that's also what the map companies are working on, towards a more intelligent and efficient system.

Cornick et al. [44] created a new map-based localization called Localized Ground-Penetrating RADAR (LGPR). As the name implied, their system mounted at the under-frame of a car sends electromagnetic pulses towards the ground and measures the unique reflection profiles from under the surface. With highly efficient process units matching the identifiers with registration data, they manage to acquire the precise location of an ego vehicle up to the speed of highway standard. Different from common automotive radar, this system uses 100 MHz to 400 MHz radar with the ability in operating under all kinds of rough environments. Not only does it achieve at least equal or better accuracy than traditional localization methods, but it also shows a great advantage in navigating water puddles or snowdrifts. It does have limitations though. When the saturation of soil reaches a certain level like 30%, the attenuation it caused for radar at their operating range could reach 10 dB/m [249]. An additional risk in the snow is the snow melting salt which attenuates the RF signal kind of severely. Although margins have been reserved to overcome these, they are still working on better characterizations on the LGPR system.

Sensible4 uses a LiDAR-based volumetric probabilistic distributions [250] approach for mapping and localization in all-weathers, which reduces the amount of information and details in the map, and instead captures the distribution of patches of the surface (surfels). They create such maps from year-long LiDAR data, therefore the common elements of the road are modeled regardless of the season.

### 7.3. Sensor solutions

All the participated sensors in ADS, LiDAR, radar and camera get their chances to carry localization in harsh conditions, and multi-sensory place recognition is proven to be more robust [251]. The radar extension of the Oxford Robotcar dataset [75] used only Navtech radar to accomplish map building and localization in rain and snow, thanks to radar's specialty in harsh weather. Wolcott and Eustice [232] developed a robust LiDAR localization using multi-resolution Gaussian mixture maps. They discretized the point cloud into grids of 2D and expand Gaussian mixture distribution on the z-axis. This helps compact point clouds into compressed 2.5D maps with parametric representations. They demonstrated the ability to navigate all kinds of road texture conditions including constructions, and during harsh weather like snow. Their tests showed a small increase of root square mean (RSM) error in normal downtown conditions compared to traditional reflectivity-based localization which was not considered a problem, they did decrease the RSM error rate by around 80% on both lateral and longitudinal sides in snowy conditions [233].

Similar to sensor fusion, extra modalities are also enlisted to help localization. Brunner et al. [235] combined visual and infrared imaging in a traditional SLAM algorithm to do the job. They evaluate the data quality from each sensor first and dispose of the bad ones which might induce errors before combining the data. The principle of introducing thermal cameras here is almost the same as discussed before, only for localization purposes here particularly. Their uniqueness is that they not only tested the modality in low visibility conditions, like dusk or sudden artificial strong light, but also tested in the presence of smoke, fire and extreme heat, which saturate the infrared cameras. There's no guarantee that the flawed data have no weight in the algorithm at all but the combination definitely reduces the error rate compared to a single sensor modality. Their method makes the best of each sensor in situations that are not seriously affected and holds back the error of those who are affected. This tactic of playing up strengths while avoiding weakness gives us the chance of resilient perceptions.

### 7.4. Lateral localization

Besides the traveling direction along the roads, vehicles' localization in the lateral direction perpendicular to the traveling motion is equally important because keeping inside the lane while cruising is an important traffic rule and one of the guarantees of safe driving. The challenges that lateral localization is facing in adverse weather conditions are that many elements used for normal localization such as the lane lines, curbs, road barriers and other landmarks, as well as the sensors to perceive them may be unavailable or under influence at the moment. In some cold places, even on a clear day, the road might be covered by accumulated snow; the noise or interference received by LiDAR or cameras during rain or snow might happen to block essential reference lane marks; therefore the AV would have a hard time to put itself in the right lane for turning or navigating through intersections in such situations.

Aldibaja et al. [252] described the general reasons for lateral localization drifting by converting map images into edge profiles to represent the road marks in a series of LiDAR signal reflectivity peaks. Accumulated snow on the roadside creates sharp intensity peaks with irregular distribution for LiDARs while wet elements from snow-rain weather leave a track of line with low reflectivity on the road. The wearings of old roads and vegetation whose branches reach into the road space also create anomalies sometimes. These confuse the LiDAR about the actual whereabouts of the lane lines and the boundaries of the road area which leads to wrong lateral movements. Aldibaja's group proposed to use Principal Component Analysis (PCA) method to extract dominant edge profile distribution patterns and eliminate the "fake" lane lines via edge profile matching [122] [234]. Also by patching the missing LiDAR elements based on leading eigenvectors (eigenroads), a reliable LiDAR profile was reconstructed. The error in the lateral movement was reduced to 15 cm with a localization accuracy of 96.4% in critical environments [252].

## 8. Planning and Control

By now, pretty much all the essential information has been collected and processed to ensure the safest and efficient decision was made by the planner. Finally, the actuators execute the commands and do the "driving". A summary of planning and control related to weather is shown in Table 4.

### 8.1. Planning

The influence of weather on global route planning might have not yet come into people's sight because it would take a long-distance field test and unpredictable weather to meet the conditions which are hard to run into. However, imagine meteorological disasters or geological disasters caused by extreme weather such as hurricanes or floods, damage the roads or pose threats to outdoor activities, the global path planning module has to adjust its original route and navigate the AV around the impacted areas. Similar to the navigation we have on our smartphones right now, map apps can alert customers if there's an accident or road work on the way and automatically optimize the route when the estimated time is enlarged due to the situation. The Civil Engineering Research Institute of Cold Region (CERI) Snow and Ice Research Team [253] is developing a wide-area information provision service in the northern part of Japan where snow is common. This service can send alerts or emails about current snowfall, visibility and forecast information to drivers' phones and provide detour options around snowstorms. For AVs, such function relies on Internet-connected devices and mass data mining, so IoT or at least Vehicle to Vehicle (V2V) is needed for tasks like this for weather conditions, just like the flood warning system previously introduced in [125].

It is worth mentioning that the winter driving pattern and road wearing prevention is an issue of concern in local route planning, as autonomous driving is a continuous refinement process: if a shuttle service AV drives the same route over and over without any variation introduced to its driving lines, the precision of its driving route would approach to a level of centimeters over time and the road would be facing a worn groove, or a perilously slippery track in winter [237].

Taş et al. [243] presented an uncertainty-aware motion planning under limited visibility which could be the result of adverse weather or occlusions. Their planner imitates human drivers' behaviors in bad driving conditions including lowering speed, preparing to yield for not-yet visible approaching vehicles, and being on the lookout for incompliant behaviors despite our own right-of-way. By assuming the worst case scenario, the motion planner is made towards robustness. Jalalmaab et al, [238] designed a model predictive controller (MPC) with time-varying safety constraints for path planning with collision avoidance. The proposed controller takes the road boundary and dynamic change of surrounding vehicles into the constraints consideration and finds the best commands of longitudinal and lateral control to navigate the AV. Although it's not designed for weather conditions, the weather is just another parameter that also possesses the time-varying dynamic feature. It can either be quantifiable perception data or road conditions data, which will be introduced in Section 9. Peng et al. [239] introduced an adaptive path planning model and tracking control method for collision avoidance and lane-changing maneuvers in rainy weather. They first use Gaussian distribution to evaluate the impact of rain on the pavement and deduce the adaptive trajectory and the followed score-based decision making and multilevel driving mode take control while maintaining safety, comfort, and efficiency. It can be seen that this kind of motion planning needs the foundation of weather classification or uncertainty level assessment. Hence, the integration of sensors and function modules is vital for omnipotent ADS.

### 8.2. Control

No doubt, the patterns of vehicle control or decision-making in adverse conditions are not going to remain exactly the same in order to adapt. A typical example: an AV is caught in a thunderstorm on the road out of blue with visibility plummeting. It is critical for the AV to make adjustments to deal with the adverse condition including slowing down, activating countermeasures against low visibility and wet road surface and so on. It is well accepted that when the road surface is wet or icy, the braking distance would be longer and the risk of rear-end collision is largely increased. According to the research of Ali Abdi Kordani et al. [240] on the vehicle braking distance at the speed of 80 km/h, the road friction coefficient of rainy, snowy and icy road surface conditions are 0.4, 0.28 and 0.18 respectively, while average dry road friction coefficient is about 0.7. While the braking distances under rainy conditions barely change much for buses or trucks considering their own high inertia, a 10-meter increase can be noticed for a sedan; The braking distance rises 28% in snowy conditions and dramatically surges 71% in icy conditions, up to 180 m for a normal sedan and 300 m for trucks and buses [254]. This startling data remind us that the behavior of an AV needs certain adjustments under adverse conditions, especially when the road surfaces are wet or frozen. The largely prolonged braking distance requires a substantially longer following distance and perhaps a lower cruising speed if necessary. A traction mode shift for certain types of cars might also be required when the road condition changes. As a result, the weather effects caused on AVs directly generate demand for behavior adjustment.

When driving in gale weather or on a bridge over a large area of water, a vehicle sustains a large force perpendicular to the vehicle's heading direction. This is called side-slip, often caused by crosswind. Crosswind stabilization normally is the job of the electronic stability control system (ESC), which stabilizes the lateral motion of the vehicle to prevent skidding [255]. However, serving as redundancy, the ESC system is initiated belatedly, using the brakes to correct the deviation from a yaw reference model when the vehicle starts to skid [256]. For ADS, it's better to have a proactive mode of steering control that has better judgment than a human driver.

Besides algorithm models, the industry is more concerned about the real performance of AV control. Risk assessment is a primary basis for motion planning and control, while weather conditions geometrically increase the risk index and the level of control difficulty. The mismatch of wheel encoder and the real vehicle motion caused by slipping and traction loss mentioned above not only poses threats to the ego vehicle's localization, but also to the successful execution completion of AV control commands [241]. Autonomous driving technology provider Sensible4 company implements a control stack with multiple sensors monitoring the wheels. The information on the actual acceleration, wheel rotation speed, and wheel angle are critical for the control module to make the right amendment and maintain accurate control [237]. To say the least, there is always a final barrier of AV control which is the remote control for vehicles of level 4 or higher where no safety pilots are present [242]. A staff would sit in the remote control center monitoring the AV's movement and data, and take over whenever it's necessary.

Last but not least, passenger comfort is not something that should be overlooked by ADS providers when it largely depends on the AV's control algorithms [257]. Unnecessary acceleration and abrupt braking should be avoided for the best riding experience for a passenger, but that's not all. Should there be any adverse driving conditions, it's the distrust and lack of confidence in the AV's behaviors that disturb people's nerves. Therefore, in order to assure the consumers, installing information display systems such as heads-up displays (HUD) which prompt the AV's upcoming movements especially under low visibility conditions can yet be regarded as a good choice for technology companies and manufacturers [244].

## 9. Auxiliary Approaches in Adverse Conditions

Sometimes technologies beyond the vicinity of an ego vehicle itself and unconventional approaches also help with weather conditions. A summary of auxiliary approaches in adverse weather solutions is shown in the first part of Table 5.

### 9.1. Road surface detection

Originally, road surface detection was designed to classify the road type like asphalt, grass, gravel, sand, etc., to better provide traction ability. On the other hand, road surface conditions' change is a direct result of weather conditions, especially wet weather, so it's natural to think of road surface detection as an auxiliary for weather classification. Although the mirage effect also happens on the surface of roads due to high temperature, it's more of a light related situation rather than physical changes to the roads, which has been discussed in Sec.5.4.

As stated in Sec.8.2, wet or slippery road surfaces pose an actual threat to traffic safety. The information on road conditions could be as important as weather conditions for a car. The dry or wet conditions can be determined in various ways besides road friction or environmental sensors [258]. Sabanovic et al. [259] build a vision-based DNN to estimate the road friction coefficient because dry, slippery, slurry and icy surfaces with decreasing friction can basically be identified as clear, rain, snow, and freezing weather correspondingly. Their algorithm detects not only the wet conditions but is able to classify the combination of wet conditions and pavement types as well. Although they didn't perform implementation on autonomous driving, it's suitable for improvement of vehicle dynamic control and anti-lock braking system (ABS) which are also critical for ADS. Panhuber et al. [260] have a successful application of road surface detection in autonomous driving. By mounting a mono camera behind the windshield and observing the spray of water or dust caused by the leading car and the bird-view of the road features in the surroundings, they determine the road surface condition with multiple classifiers. The wet or dry classification accuracy is 86% for autonomous driving and it's a great asset of the weather classification module in ADS.

The road surface detection can also be performed in a way that is unexpected for us: audio. The sounds of vehicle speed, tire-surface interaction, and noise under different road conditions or different levels of wetness could be unique, so it's reasonable for Abdic et al. [261] to train a deep learning network with over 780,000 bins of audio, including low speed when sounds are weak, even at 0 speed because it can detect the sound made by other driving-by vehicles. There is a concern that the vehicle type or tire type may affect the universality of such a method and the uncertain degree of difficulty of the installation of sound collecting devices on AVs. All these tell us there is still a lot to explore on the car or the road to find something that can be of help in adverse weather conditions.

### 9.2. Roadside Units

Roadside LiDAR, also from a different point of view than the top of a car, has already started to participate in autonomous driving. Similar to bird-view units, a LiDAR installed on a roadside unit (RSU) mounted on a pole also has a wide angle of view and doesn't suffer from the water screen on rainy days or the snow whirl in snowy weather caused by surrounding vehicles. A slightly downward pointing angle also dodges some direct sunlight. Wu et al. [262] have tested the roadside LiDAR's vehicle detection performance in windy and snowy conditions, and improved the data processing with background filtering and object clustering. If there is a connected system available, then AV can utilize the data captured by the roadside LiDAR and make better planning.

Sometimes, when weather conditions are too bad for any sensor to perform at a reliable status like dense fog, the data provided by roadside LiDAR could serve as redundancy to collaborate the decision made by the ADS on AV and reduce accidents in important locations. Tian [263] used roadside LiDAR to identify weather conditions. They recorded the LiDAR point cloud pattern characteristics of each kind of weather including rain, snow, and wind, and quantified the impact of each weather so as to differentiate the weather based on the standard deviation of their detection distance offsets. Their data don't include fog conditions which could be rare in Nevada, and it reveals the geographic limitation of roadside LiDAR. Vargas Rivero et al. [264] placed a LiDAR at a fixed location pointing at the parking lot asphalt for 9 months and classified multiple weather according to weighted k-nearest neighbors algorithm (k-NN) scores derived from several LiDAR parameters, including number of detections in the maximum of the histogram, mean detection distance in the x-direction, standard deviation of the detection distance in the x-direction, number of second echo detections and mean value of the echo-pulse width (EPW). The accuracy F-score normally stays above 90% for clear, fog, rain and snow conditions.

In the future, with a connected vehicle system or a V2X system, RSU can at least provide weather classification information and perception data from its own location to nearby vehicles and that's already better than AV alone.

### 9.3. Others

Inclement weather conditions bring challenges and impacts to human drivers in the same way before AV became popular. Consequently, some of the solutions designed for building up human drivers' arsenals against weather can still be of use for today's AV. We know that visibility drops when it's raining or snowing, especially at night. Back when digitally removing rain and snow streaks in images was trending in computer vision, Charette et al. [265] presented a smart automotive headlight that can see through rain and snow by deactivating the light rays that intersect with raindrop or snowflakes particles. They tracked and avoided illuminations on the raindrops with only very few losses on light throughput and successfully achieved an obvious visibility improvement. Given that what on-board cameras now see is not that different than what human drivers saw back then, such smart headlights should be able to improve visibility for ADS as well. Donati et al. [266] even fabricated a smart headlight embedded with a charged-coupled (CCD) camera and a LiDAR that can intelligently switch off high beams when crossing and being able to detect human-sized

**Table 5**
Summary of adverse weather auxiliary and future research

| Road Surface | RSU | Head Lights | Human Detection | Tracking | V2V | V2I | Aerial View |
|---|---|---|---|---|---|---|---|
| [259] | [262] | [265] | [272] | [275] | [276] | [281] | [284] |
| [260] | [264] | [266] | [270] | [274] | [277] | [276] | [285] |
| [261] | [263] |  | [269] | [273] | [125] | [282] | [286] |
| [258] |  |  | [271] |  | [278] | [283] |  |
|  |  |  |  |  | [279] |  |  |
|  |  |  |  |  | [280] |  |  |

objects at 20 m distance with a recognition rate of 85% in the meantime.

There is an interesting platform named OpenXC [267], an Application Programming Interface (API) that combines open source software and hardware to extend vehicles with custom applications and modules. Most of the applications are based on Android that can read and translate car metrics, and most importantly, they are connected service integration ready. Take two examples related to adverse conditions. The Nighttime Forward Collision Warning (Night Vision) project uses a standard USB webcam to capture the edges of an object to do object detection, and helps avoid animal or obstacle collisions. The Brake Distance Tracking project measures the distance between vehicles with a SICK DMT-2 LiDAR [268] sensor and warns drivers when they are approaching other vehicles with too high momentum. These setups seem simple, but they all run on Android platforms such as a tablet with access to sensor data and real-time vehicle data which are of high compatibility and integration flexibility.

Some other efforts have also been made on restoring or improving the performance of ADS' basic functions like human/pedestrian detection and vehicle tracking. Recognition of the particular micro-Doppler spectra [269] and multi-layer deep learning approaches [270] are used in pedestrian detection tasks in bad weather. Thermal datasets specifically targeting pedestrians [271] or large scale simulation dataset [272] are also being established to make sure that ADS can complete this essential job with the interruption of weather. While conventional Gaussian mixture probability hypothesis density filter based tracker is being utilized in deep learning vehicle tracking framework [273] to improve the performance, radar tracking [274] and color-based vision lane tracking system on unmarked roads [275] also show the robustness and can help ADS maintain functionalities while in adverse conditions even if they were not particularly designed for those conditions.

### 10. Tools

#### 10.1. Data sets

Autonomous driving research can't be done without datasets. Many features used in object detection tasks need to be extracted from datasets and almost every algorithm needs to be tested and validated in datasets. In order to better tackle the adverse weather conditions in autonomous driving, it's essential to have enough weather conditions covering each kind of weather in datasets. Unfortunately, most of the datasets commonly used for training do not contain too many conditions different from clear weather. Some famous datasets that were collected in tropical areas like nuScenes [113] contain some rain conditions in Singapore, A*3D [295] has rain conditions at night, and ApolloScape [297] includes some strong light and shadow conditions. A summary of the weather conditions coverage and the sensors used for collection in each dataset is shown in Table 6.

**Table 6**
Coverage of weather conditions in common autonomous driving datasets

| Dataset | Synthesis | Rain | Fog/Haze/Smog | Snow | Strong Light/Contamination | Night | Sensors |
|---|---|---|---|---|---|---|---|
| LIBRE [2] | - | ✓ | ✓ | - | ✓Strong light | - | 10 LiDARs, Camera, IMU, GNSS, CAN, 360° 4K cam, Event cam, Infrared cam |
| Foggy Cityscape [193] | ✓ | - | ✓ | - | - | - | - |
| CADCD [58] | - | - | - | ✓ | - | - | 1 LiDAR, 8 Cameras, GNSS, IMU |
| Berkley DeepDrive [287] | - | ✓ | ✓ | ✓ | - | ✓ | Cameras |
| Mapillary [288] | - | ✓ | ✓ | ✓ | - | ✓ | Mobile phones, Tablets, Action cameras, Professional capturing rigs |
| EuroCity [289] | - | ✓ | ✓ | ✓ | - | ✓ | 2 Cameras |
| Oxford RobotCar [290] | - | ✓ | - | ✓ | - | ✓ | 3 LiDARs, 3 Cameras, Stereo cam, GPS (Radar extension: 360°radar) |
| nuScenes [113] | - | ✓ | - | - | - | ✓ | 1 LiDAR, 6 Cameras, 5 Radars, GNSS, IMU |
| D2-City [291] | - | ✓ | ✓ | ✓ | ✓Contami-nation | - | Dashcams |
| DDD17 [292] | - | ✓ | - | - | - | ✓ | Dynamic and active-pixel vision Camera |
| Argoverse [293] | - | ✓ | - | - | - | ✓ | 2 LiDARs, 7 Cameras ring, 2 Stereo cams, GNSS |
| Waymo Open [294] | - | ✓ | - | - | - | ✓ | 5 LiDARs, 5 Cameras |
| A*3D [295] | - | ✓ | - | - | - | ✓ | 1 LiDAR, 2 Cameras |
| Snowy Driving [296] | - | - | - | ✓ | - | - | Dashcams |
| ApolloScape [297] | - | ✓ | - | - | ✓Strong light | ✓ | 2 LiDARs, Depth Images, GPS/IMU |
| SYNTHIA [298] | ✓ | - | - | ✓ | - | - | - |
| P.F.B [299] | ✓ | ✓ | - | ✓ | - | ✓ | - |
| ALSD [272] | ✓ | ✓ | - | ✓ | - | ✓ | - |
| ACDC [300] | - | ✓ | ✓ | ✓ | - | ✓ | 1 Camera |
| NCLT [301] | - | - | - | ✓ | - | - | 2 LiDARs, 1 Camera, GPS, IMU |
| 4Seasons [226] | - | ✓ | - | - | - | ✓ | 1 Stereo Camera, GNSS, IMU |
| Raincouver [302] | - | ✓ | - | - | - | ✓ | Dashcam |
| WildDash [303] | - | ✓ | ✓ | ✓ | - | ✓ | Cameras |
| KAIST multispectral [304] | - | - | - | - | ✓Strong light | ✓ | 1 LiDAR, 2 Cameras, 1 Thermal (infrared) cam, IMU, GNSS |
| DENSE [98] | - | ✓ | ✓ | ✓ | - | ✓ | 1 LiDAR, Stereo Camera, Gated Camera, FIR Camera, Weather Station |
| A2D2 [305] | - | ✓ | - | - | - | - | 5 LiDARs, 6 Cameras, GPS, IMU |
| SoilingNet [176] | - | - | - | - | ✓Contami-nation | - | Cameras |
| Radiate [306] | - | ✓ | ✓ | ✓ | - | ✓ | 1 LiDAR, 1 stereo camera, 360°radar, GPS |
| EU [307] | - | - | - | ✓ | - | ✓ | 4 LiDARs, 2 stereo cameras, 2 fish-eye cameras, radar, RTK GPS, IMU |
| HSI-Drive [308] | - | ✓ | ✓ | - | - | ✓ | 1 Photonfocus 25-band hyperspectral camera |
| WADS [309] [310] | - | ✓ | - | ✓ | - | ✓ | 3 LiDARs, 1 camera, 1 NIR camera, 1 LWIR camera GNSS, IMU, +1550 nm guest LiDAR |
| Boreas [311] | - | ✓ | - | ✓ | - | ✓ | 1 LiDAR, 1 camera, 1 360° radar, GNSS-INS |

**Table 7**
Weather conditions and sensors support in simulators

| Simulators | Weather conditions | | | | | | Sensor support | | | | | | |
|---|---|---|---|---|---|---|---|---|---|---|---|---|---|
| | Adjustable | Rain | Fog | Snow | Light/Time of day | Contamination (Dust, leaf) | LiDAR | Camera | Thermal Camera | Radar | GNSS/GPS | Ultrasonic | V2X |
| CARLA [312] | ✓ | ✓ | ✓ | - | ✓ | - | ✓ | ✓ | - | ✓ | ✓ | ✓ | - |
| LG SVL [313] | ✓ | ✓ | ✓ | - | ✓ | - | ✓ | ✓ | - | ✓ | ✓ | ✓ | - |
| dSPACE [314] | ✓ | ✓ | ✓ | ✓ | ✓ | - | ✓ | ✓ | - | ✓ | ✓ | ✓ | ✓ |
| CarSim [315] | - | ✓ | - | ✓ | ✓ | - | - | ✓ | - | ✓ | ✓ | ✓ | - |
| TASS PreScan [316] | - | ✓ | ✓ | ✓ | ✓ | - | ✓ | ✓ | - | ✓ | ✓ | ✓ | ✓ |
| AirSim [317] | ✓ | ✓ | ✓ | ✓ | ✓ | ✓ | ✓ | ✓ | - | - | ✓ | - | - |
| PTV Vissim [318] | - | ✓ | ✓ | ✓ | - | - | Need to integrate with other platforms | | | | | | |

Researchers collected weather data that are common in their area of living or used simulation [272] to build their own weather datasets. The University of Michigan collected four-season LiDAR data using a Segway robot on the campus at an early stage [301]. Pitropov et al. [58] presented the first AV dataset that focuses on snow conditions specifically, called the Canadian adverse driving conditions (CADC) dataset. The variety of winter was collected by 8 cameras and LiDAR and GNSS+INS in Waterloo, Canada. Their LiDAR was also modified by the de-noising method [160]. The large amount of snow enables researchers to test object detection, localization and mapping in all kinds of snow conditions, which is hard to realize in artificial environments. Oxford RobotCar [290] is among the early datasets that put weights on adverse conditions including heavy rain, snow, direct sunlight and night, even road and building works. Sakaridis et al. [193] applied foggy synthesis on the Cityscape dataset [327] and generated Foggy Cityscapes with over 20000 clear-weather images, which is wildly used in the de-hazing task. The same team later introduced ACDC [300], the Adverse Conditions Dataset with Correspondences for training and testing semantic segmentation methods on adverse visual conditions. It covers the visual domain and contains high-quality fine pixel-level semantic annotated fog, nighttime, rain and snow images. Zheng et al. [328] uploaded the IUPUI Driving Video/Image Benchmark to the Consumer Digital Video Library, containing sample views from in-car cameras under different illumination and road conditions when public safety vehicles are on patrol and responding to disasters. Conditions cover snow, rain, direct light, dim lit conditions, sunny facing the sun, shadow, night, and their caused phenomena such as wet roads, glass reflection, glass icing, raining and dirty windshields, moving wipers, etc. It's the unremitting effort of research on collecting data on cold days and dangerous driving conditions that gives us the opportunity to push autonomous driving in adverse conditions further to the next level.

**Table 8**
Experimental weather facilities across the world

| Experimental facilities | Adjustable | Rain | Fog | Snow | Light/ Time of day | Contamination (Dust, leaf) | Location | Length | Lanes |
|---|---|---|---|---|---|---|---|---|---|
| JARI Special Environment Proof Ground [319] | ✓ | ✓ | ✓ | - | ✓ | - | Ibaraki, Japan | 200 m | 3 |
| VTTI Smart roads [320] | ✓ | ✓ | ✓ | ✓ | ✓ | - | Virginia, US | 800 m | 2 |
| DENSO [321] | ✓ | ✓ | - | - | ✓ | - | Aichi, Japan | 200 m | 10 m wide |
| Center for Road Weather Proving Ground [322] | ✓ | ✓ | ✓ | ✓ | ✓ | ✓ | Yeoncheon, Korea | 600 m | 4 |
| Laboratoire Régional des Ponts et Chaussées, Clermont-Ferrand [323] | ✓ | ✓ | ✓ | - | - | - | CEREMA, France | 31 m | 2 |
| NIED Cryospheric Environment Simulator [324] | ✓ | ✓ | ✓ | - | - | - | Yamagata, Japan | N/A | N/A |
| CATARC Proving Ground [325] | ✓ | ✓ | ✓ | ✓ | ✓ | - | Yancheng, China | 60 km track | 2 or more |
| CERI Tomakomai Cold Region Test Road [326] | ✓ | - | - | ✓ | - | - | Hokkaido, Japan | 21 hectare ring track | 2 |

### 10.2. Simulators and Experimental facilities

The rapid developments of autonomous driving especially in adverse weather conditions benefit a lot from the availability of simulation platforms and experimental facilities like fog chambers or test roads. Virtual platforms like the well-known CARLA [312] simulator, as shown in Fig.19, enable researchers to construct custom-designed complex road environments and non-ego participants with infinite scenarios where it would be extremely hard and costly in real field experiments. Moreover, for weather conditions, the appearing of each kind of weather especially season-related or extreme climates related is not on call at all times. For example, it's impossible for tropical areas to have the opportunity to do snow tests; and natural rain showers might not be long enough to collect experimental data. Most importantly, adverse conditions are usually dangerous for driving and subjects always face safety threats in normal field tests, while absolute zero risks are something that simulators can guarantee. Simulators provide a great platform for research in autonomous driving in adverse weather conditions anytime and anywhere. In recent years, various simulation platforms including open source and closed source software have developed adjustable weather conditions and 'time of day' plug-ins. Thus, people can test their ADS modalities against rain, snow, fog with different precipitation rates, and strong light in simulators before taking it to the outside [329]. Table 7 lists the weather conditions and sensors supported in some common autonomous driving simulators.

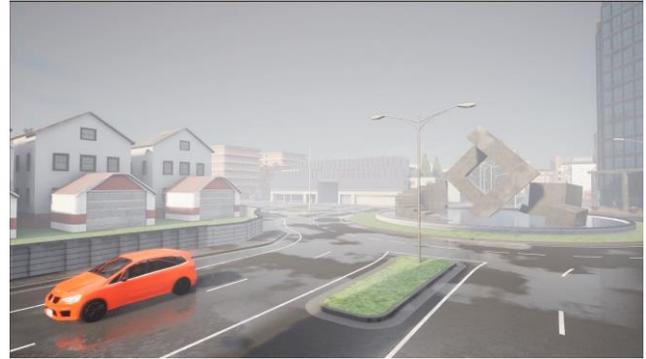

**Figure 19:** A scenario of a town with wet road surface in fog weather in CARLA simulator.

On the other hand, laboratory environments can also replace real field tests with control. Considering the limitation on test fields and safety hazards to the surrounding people or facilities, an enclosed artificial track or chamber with rain/snow making machines offers almost the same environmental conditions with the advantage of controllable precipitation rates and low risks. To say the least, even with enough resources and the perfect weather conditions that fit the researchers' needs, current legislation can hardly support AV tests with autonomy equal to or larger than level 4 in adverse conditions and inclement weather. Table 8 lists out some renowned test facilities with some details about the chambers or tracks.

## 11. Trends, Limitations & Future Research

This section will discuss the trend of current adverse weather research, the limitations we are facing along the road, and potential research directions and focus in the future. A summary of the literature covered regarding future research is listed in the second part of Table 9.

### 11.1. Social significance

In all automated vehicle projects, it is critically important to recognize and resolve a broad range of legal, regulatory, and liability issues. Fortunately, legislation around the world is keeping well pace with current research trends and mostly allows up to level 4 operating (conditions apply). The Mcity project as shown in Fig.1(b), is a good role model of benefiting from Michigan being one of the first States that allow level 4 automated vehicles to be tested on open roads [330]. Their operation protocol requires that the person "operating the vehicle" be able to take control of the vehicle's movements in case of an emergency. Alternatively, if the person cannot take control, then the vehicle must "be capable of achieving a minimal risk condition" [3]. That's why the Mcity shuttles are still equipped with a conductor on board who has override control even though there's absolutely no steering wheel or any other traditional maneuver installation whatsoever. Another major power in automotive, Japan, also revised related legislation in 2020 allowing level 3 or above operations [331], and level 4 autonomous shuttles from Toyota were deployed to ferry athletes during the 2020 Tokyo Olympic Games. However, one of these autonomous buses hit a visually impaired Paralympian [332], and this is the very reason for further autonomous driving research. It's still exceptionally helpful that authorities create plenty of room for ADS research by exempting the absolute requirement of human driver presence inside the vehicle.

Social surveys show that people who have had experiences with AVs, either rode with one or saw one operating in real life, have a more than 75% rate of satisfaction and recognition, and 60% more interest in this subject [333], which indicates a positive perspective in the further development of Autonomous Vehicles. The results from South Korea Yonsei University's mathematical model demonstrated that trust and perceived usefulness are the core determining factors of the intention of AV uses [334]. One major source of said trust is technical competence which means the pending challenge in adverse weather plays a heavy role in the public acceptance of AVs.

### 11.2. Trends
#### 11.2.1. Toward advanced sensor fusion

To some extent, the tendency of autonomous driving is affected by market preference and product attractiveness, because the commercial and economic value is part of the motive force of ADS research. This is why after using mere radar as the main core sensor for years [335], Tesla later then announced that they started transitioning to a camera-based pure vision Autopilot system [336]. However, this route does not exactly fit our expectations on better tackling adverse weather conditions, as the deficiency of the sole force of each ADS sensor alone in weather is well established. Each sensor has its own strength against particular problems, such as the lower signal attenuation radar possesses over LiDAR in rain conditions previously mentioned in Sec.3.2. It's of ADS's best interest to make the best of each sensor's superiority where happens to be others' deficiency. As summarized in Sec.4, sensor fusion, the combining force of several sensors is still the most reliable way to build a robust modality that is agnostic to weather.

While some trying to enlist every help available by appending all kinds of specialized sensors such as the weather station, the sensible way is to pick out the best-performed combination of necessaries and maintain a feasible economic cost and computational cost in the meantime. With the development of electric cars moving towards larger battery capacity and endurance, it's reasonable to believe that more sensors can be supported on one AV. Additionally, with electric cars manufacturing and assemblies maturing out of the traditional auto industry, more than one type of sensors, such as directional cameras or LiDARs, are going inside the AV's body structure like the grills, and become build-in sensor sets which is a huge step further than the sensor rack on top of a traditional car with a storage battery module in the trunk. Hesai launched a new short-range blind spot LiDAR, QT128 [337], at the beginning of 2022, which is designed to be installed at the front-end of a car to deliver environmental details with calibrated reflectivity values within the ego-vehicle's about 20 m proximity. As many as the fusion combinations have been tried, the choices haven't been exhausted yet. The studies on LiDAR dominant fusion modalities and thermal-camera-included modalities have yet to joined forces very closely. Future benchmarking and evaluations at a comprehensive level on advanced sensor fusions should provide the community with one or more well-recognized modalities that can make us feel safe when cruising in weather.

Besides the common sensor options, there are also new types of LiDARs and cameras on the table right now that have started to lead the major research trend among many technology companies around the world. For example, solid-state LiDAR and MEMS LiDAR who largely depend on the semi-conductor industry. Unlike traditional rotating LiDARs, solid-state LiDAR needs beam steering to tune the laser direction and one of the popular ways is through Optical Phased Array (OPA) platform. Thermal optics tuning is currently the dominant method as the thermo-optic coefficients of the two major materials, Si and SiN, have a difference of over an order of magnitude [338]. Tunability of thermal tuning is somehow limited, whereas wavelength tuning could achieve a tunability of a couple of dozen degrees per 100 nm change around 1550 nm laser wavelength [339]. There are also other cutting-edge beam steering techniques, such as metasurfaces [340], starting to emerge these years.

FMCW (Frequency Modulated Continuous Wave) LiDAR is a technology using continuous waves to do coherent detection which also caught a great deal of attention recently. Unlike traditional LiDAR using amplitude modulation (AM) approach, FMCW LiDAR emits a continuous laser beam to measure the change in frequency of the waveform as it reflects off of an object, which gives it the ability to see as far as more than 300 meters and measuring the instantaneous velocity based on Doppler shift [341]. The longer detection range and the accurate velocity sensing help identify a pertinent high moving subject at a distance and give the AV enough time to react. Even though the difference could be as small as a fraction of a second, it would still make a huge difference for a heavy vehicle, like a bus or cargo truck, given its enormous inertia. Most of all, lights that do not match the FMCW LiDAR's local oscillator are not detected, which provides an immunity to interference from solar light and the cross-talk between other AVs or even the ego vehicle's previous signal itself. This strong point brings the attention of this kind of LiDAR to solving our weather problems but still with limitations which will be discussed in Sec.11.3.1.

Similar to the Navtech radar [76] we mentioned before that has lifted the upper bound of sensor use, cameras that are suitable candidates for advanced sensor fusions won't be limited to plain traditional cameras either. Instead, professional cameras have started to be deployed for data collection and replaced conventional imaging devices. Some examples are shown below:

A stereo camera has two or more lenses with a separated image sensor for each lens, which provides the ability to capture 3-D images, just like human binocular vision.

Thermal camera uses infrared radiation to create images. Far-infrared (FIR) cameras operate at 8 - 12 $\mu m$ and can see heat sources, while near-infrared (NIR) camera normally operates around 700 - 1400 nm and can penetrate what visible light could not, like haze, light fog and smoke [342].

Event camera, such as the dynamic vision sensor (DVS), does not capture images using a shutter as conventional cameras do, but individual and asynchronous pixels that report any brightness changes [343]. Event camera offers a very high dynamic range and no motion blur, but traditional vision algorithms do not apply to asynchronous events output, so the application on cars normally would require additional algorithms.

For the OpenCV OAK-D AI cameras, the fusion even happens before our definition of sensor fusion, for this type of camera is consisted of a high-resolution RGB camera, a stereo pair, and an Intel Myriad X Visual Processing Unit (VPU), which can produce a depth map with sophisticated neural networks for visual perception [344]. This train of thought also brings up the next stage of the trends of ADS development.

*11.2.2. Toward sophisticated networks*

Successful sensor fusions rely on the strength of each of their element, but it would be hard to release their full potential on perception without sophisticated algorithms and machine learning techniques helping with the processing of fusion data, as presented in [345]. Given LiDAR's particular status in ADS perception and how it's affected by weather effects, it's realistic to help LiDAR to maintain its original performance level when affected, by such as denoising and multi-echo methods, rather than improving too much. Computer vision is another story. Being the sensor that holds an absolute place in ADS, camera's perception enhancement enjoys the deepest research on algorithms and machine learning. The training on all the weather effects will be conducted to better fight the bad conditions someday, no matter how hard the dataset is to acquire because artificially creating the equivalent effects of certain weather is starting to dominate this field of research and proven useful, as introduced in Sec.5. Not only the difference from the real effect such as the contamination on camera lens is down to almost unnoticeable, the efficiency is even higher with controllable parameters.

Also, with the rapid development of AI technologies in recent years, it's not possible to get around the new methods of machine learning in ADS applications, including adverse weather solutions. For instance, the active learning of Deep Neural Network (DNN) by NVIDIA DRIVE [346]. Active learning starts with a trained DNN on labeled data, then sorts through the unlabeled data and selects the frames that it doesn't recognize, which would be then sent to human annotators for labeling and added to the training pool, and completes the learning loop. In a nighttime scenario where raindrops blur the camera lens and make it difficult to detect pedestrians with umbrellas and bicycles, active learning is proven to have over 3 times better accuracy and be able to avoid false positive detection. Other burgeoning machine learning methods such as transfer learning and federated learning could also be very effective on robust AI infrastructure, which is still left to be explored.

*11.2.3. Toward V2X and IoT*

The Vehicle-to-Everything system is an unignorable part of the development of Intelligent Transportation System (ITS) and ADS. Regardless of the proficiency of well-experienced drivers, accidents still happen due to asymmetric road information among drivers, and short reaction time. V2X system broadens the range of information gathering from one car's perception to the perception of almost every element on the road. The 'Everything', or X here, can refer to other vehicles, roadside infrastructures, and even pedestrians. To have pedestrians being included in the V2X or further Internet of Things (IoT) system, advanced wearable devices or universal smartphone technical support is needed, which is out of the scope of this survey. V2X has two major branches: Vehicle to Vehicle (V2V), and Vehicle to Infrastructure (V2I).

The core of V2V is information sharing among connected vehicles (CV), which eliminates the problem of information asymmetry from the bottom root. Under normal conditions, visual blockage like a truck in front of the ego vehicle or vehicles beyond direct line of sight due to terrain or intersections is a high risk of accidents. With V2V technology, the ego vehicle gains the ability to acquire the perception data and position information from another car whose view is out of its own reach at the moment [280], thus the driver or the ego vehicle can avoid accidents by making proper decisions and adjusting behaviors based on the additional information. In terms of harsh weather, vehicles that first experience or perceive the presence of weather conditions or the change on the road surface conditions can make weather assessments before other vehicles reach this location and then relay the perception data or assessment results to other vehicles to alert them about the dangers. If the adverse conditions cause traffic congestion or accidents, latter vehicles can plan a new route according to the information gathered by fore CVs in real time to improve efficiency and safety in intersections and work zones [276]. With the V2V technology, truck platooning on highways or docks areas is about to become the first mature application of ADS. According to the evaluation of M.Ahmed et al. [277] on the driver's perception of CV in Wyoming, almost 90% of the human test drivers found that the front collision warning and rerouting function are very useful as a benefit of the improved road condition information. As a result, it's safe to say that the improvement that V2V could bring to ADS is undoubted.

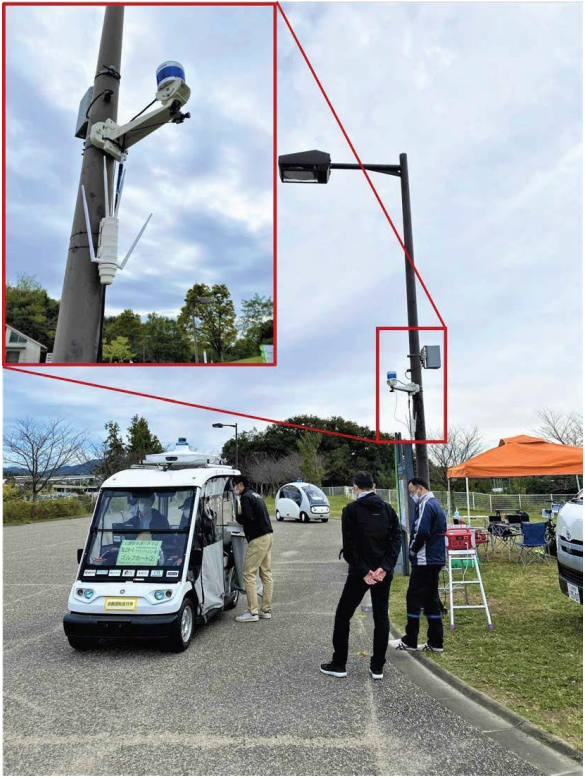

**Figure 20**: V2I infrastructures at Morikoro park in Nagoya, Japan. A self-driving vehicle interacting with the road side units (RSU). Inset: A LiDAR deployment on a pole. Images courtesy of Mr. Abraham Monrroy-Cano of Perception Engine Inc.

V2I (Fig.20) on the other hand is not as fluid as V2V, but still offers a great deal of potential in intelligent transportation. The advantages of roadside sensors like the birdview LiDAR are covered in the previous section. Besides providing alerts in specific road segments when being connected to weather information services or having its own weather perception abilities [281], V2I also provides the possibility of multi-image-based de-weathering without having to involve sophisticated neural networks discussed in Section 5. With the image of a certain scene in clear weather being captured and stored in infrastructures in advance, a 3-D model without the disturbance from weather (rain, fog, snow, etc.) can be easily reconstructed and fed to nearby vehicles to help them safely navigate under low visibility and incomplete road information.

Being a part of the future, some V2X technology has been put into use to help with adverse weather problems. Jung et al. [278] developed an ADS with V2X communication aid which is comprised of beyond-line-of-sight perception and extended planning. If we consider V2X as an extra sensor, then such a system is just like a new fusion modality. Ego vehicle's own sensors and other vehicles' sensors and roadside sensors facilitate the data input of perception including those beyond the line of sight. Route, velocity and real-time traffic information obtained by V2I communication are integrated to extend route planning with an optimal solution. Such modality approaches fully autonomous driving to a great extent but is still limited by the scale of the CV network and connected infrastructure's coverage.

Like perception methods, V2X's architecture is also being improved to better deal with weather conditions. Barrachina et al. [282] proposed a V2X-d architecture, which combines V2V and V2I together to overcome each of their own deficiency such as V2V's limited horizontal view and the vulnerability of traditional surveillance cameras on roadside infrastructures. This architecture allows vehicle density estimation in urban areas under all weather conditions which is the ideal way of utilizing the V2X system in autonomous driving. Vaidya et al. [283] used roadside infrastructures as the nearby processing and storage server of the Edge Cloud and deployed Fuzzy inference system as a road weather hazard assessment method. They took the road surface conditions and surface temperatures gathered from both roadside environmental sensors and CVs as the linguistic variables and the Fuzzy inference system outputs consequent slipperiness based on IF-THEN and AND-type and OR-type rules. Overall 10 fuzzy rules have been established. Take one as an example: IF the surface condition is 'ice warning' AND surface temperature is 'low' OR 'very low', THEN slipperiness is 'very slippery'. Such results are being processed locally in a segmented network and distributed to nearby CVs. The pressure of computing concentration, the processing time and the latency of CV communication are all largely reduced.

Take another step over V2X, it would be the era of IoT. Onesimu [279] et al. proposed an IoT based intelligent accident avoidance system for adverse weather and road conditions. They used a dataset collected by Hjelkrem and Ryeng [347] which contains not only weather conditions but also vehicle's speed and weight information which is being used to calculate the proper speed limit of the vehicle under the current condition to avoid collisions. They implemented a Naïve Bayes classifier in the training process to predict the chance of accidents in the network and use Blynk [348] mobile app to do the hardware control and simulation. This app will display a message containing an accident warning or weather information like 'Accident ahead, Sandstorm' on the back window of a car and this message will be received by the following car who would act accordingly to prevent accidents from happening, and potentially be passed along.

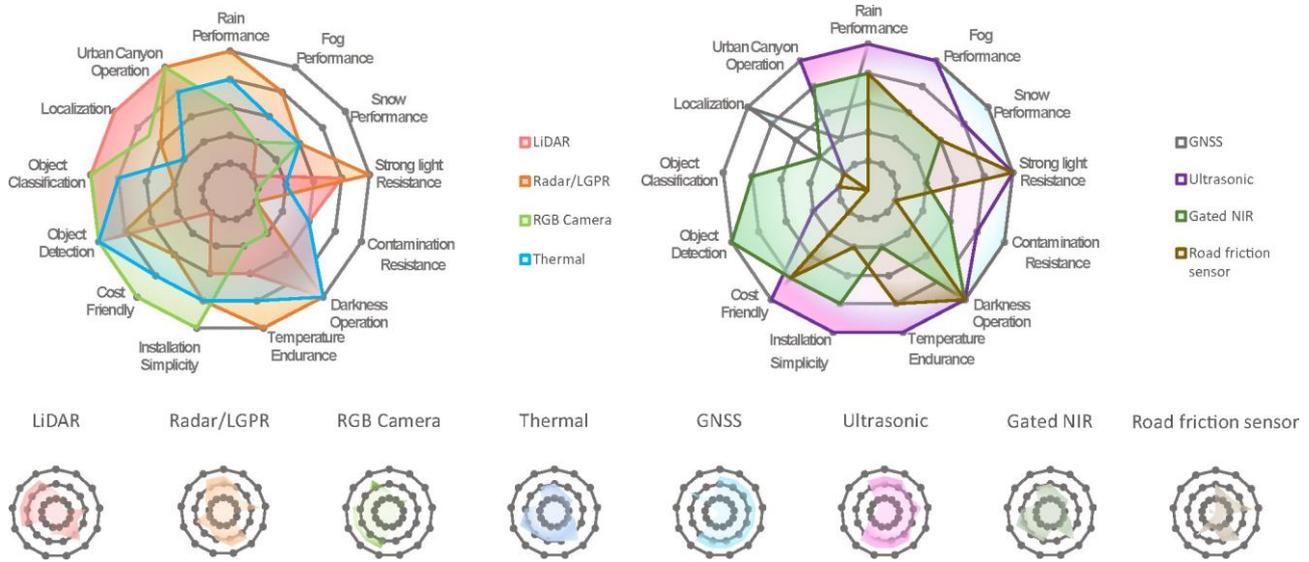

**Figure 21:** Sensor performance and characteristics radar map.

Laux et al. [349] developed one of the first complete open source vehicular networking experimental platforms in 2016 that supports most of features of the European wireless standards at the time (i.e. ETSI ITS-G5), OpenC2X. Simply a small Linux PC inside the car and the GPS receiver and antennae for wireless communication on the rooftop completes the hardware for one car in a V2X network. With its open source advantage and high compatibility with car sensors such as OBD-II for speed information, new features and updates are easily developed which is a huge contribution of OpenC2X to the V2X community. Currently, V2X and IoT as global platforms in the Intelligent Transportation System field are solving the safety problem in ADS one at a time, so there's no doubt that with profuse weather and road data, the reliability and versatility of vehicular networks will start a new page for autonomous driving.

But of course, all the features just mentioned would require real-time video (rapid image) sharing, or at least high-volume data transmission among infrastructures, vehicles and electronic devices. That's why the large volume LiDAR point cloud data need to be compressed for V2X transmission [350], and also the reason why the telecom community is working on the V2X communication methods towards richer bandwidth and lower latency such as the fifth-generation wireless technology, i.e. 5G [276] to transmit among vehicles and servers. Wi-Fi 6 (2.4 GHz and 5 GHz), based on the IEEE 802.11ax standard [351], is currently considered a well experienced IoT solution, which we could see in our daily lives such as routers and smart appliances. Qorvo for example has started the exploration of enabling a Wi-Fi 6 V2X link in the Telematics Control Unit (TCU) and antenna, and the expansion to Wi-Fi 6E (6 GHz spectrum), a critical band to establish reliable links between vehicles and their surroundings [352].

Several cities in the world, like Ann Arbor, Michigan [353]; Barcelona, Spain [354]; and Guangzhou, China [355] have initiated their smart city projects, where thousands of roadside LiDARs and sensors would be installed on city infrastructures and form a huge local connected system. It can be imagined that with developed weather perception sensors by the roadside, and every participating AV on the road gradually forming the prototype of V2X and IoT, the adverse weather conundrum would be much easier to deal with.

### 11.3. Limitations

As stated in Sec.3, the degradation of perception is the main limitation of ADS sensors in adverse weather conditions. Hardware limitations including temperature and humidity endurance are something barely analyzed systematically but shouldn't be ignored. We summarize a radar chart to show the strengths and weaknesses of each sensor in adverse conditions partially based on Table 1, as shown in Fig. 21.

### 11.3.1. 1550 nm LiDAR

Currently, the majority of the market use 905 nm wavelength LiDAR deployment. However, LiDAR upgrade has always been one of the focuses of the research community. Kutila et al. [123] raised in 2018 using 1550 nm LiDAR to overcome fog conditions because higher optical power is allowed to emit here at this wavelength. Before we determine the feasibility of this, it's necessary to bring up the two critical design considerations in LiDAR selection: eye safety and ambient suppression. Most civilian or commercial LiDARs are used in an environment where human eyes are exposed, so the infrared laser of LiDAR must not exceed the maximum permissible exposure (MPE) or cause any damage to retinas, according to the international laser product safety standard (IEC 60825-1:2014) class 1 [356]. Therefore, the selection of laser wavelength is pretty much narrowed down to two choices: 800 nm - 1000 nm and 1300 nm - 1600 nm. That's why current LiDARs made for AVs have the selection of 850 nm [357], 905 nm, and 1550 nm [358] wavelengths (see [2] for a list of other LiDARs and their respective wavelengths), and they also all fall into the window of low solar irradiance, which helps on suppressing the ambient light for the signal receiver with a lower SNR [338]. We also plot a water extinction coefficient chart as shown in Fig.22 in order to show that the 1550 nm wavelength is more likely to be absorbed by water. The extinction coefficient $\alpha_{water}$ is also known as the Lambert absorption coefficient, which is acquired from:

$$\alpha(\lambda) = 4\pi k(\lambda)/\lambda \quad (6)$$

in which $\lambda$ is wavelength and $k(\lambda)$ is the extinction coefficient of water at 25 °C. The detail of the acquisition of $k(\lambda)$ can be found in [359]. Therefore, a 1550 nm laser can be largely absorbed in the crystalline lens or the vitreous body of an eye so more energy is allowed than 905 nm, which seems to be a good thing considering the power attenuation predicament in weather [360].

However, based on the research of J. Wojtanowski et al. [361] on the comparison of 905 nm and 1550 nm performance deterioration due to adverse environmental conditions, 905 nm reaches two times further than 1550 nm in a rain rate of 25 mm/hr. Light propagation at 1550 nm might suffer less attenuation than at shorter wavelengths, but Kim et al. [362] suggested that this rule only applies to haze (visibility > 2 km), while in fog (visibility < 500 m) the attenuation is independent of wavelength and 905 nm still measures 60% longer than 1550 nm. What's more, 1550 nm waves have approximately 97% worse reflectance in snow compared to 905 nm [363]. Less interference from snow doesn't make up for the insufficiency in the original job of object detection.

FMCW LiDAR manufacturers like Aurora [366] and Aeva [367] are using 1550 nm wavelength in solid-state which is the common method nowadays. Despite its obvious advantages in filtering strong-light and anti-interference, 1550 nm wavelength's ineffectiveness in dealing with water was basically proven in the discussion above. Some promotion articles may argue that FMCW LiDAR handles the water droplets on the emitter well and can easily filter out the raindrops or snowflakes based on its velocity detecting ability [368], but the large signal attenuation that comes with the wavelength itself is omitted here and hardly any field tests of FMCW LiDAR in adverse weather conditions can be found at this stage. Notwithstanding 1550 nm's potential in further solid-state LiDAR development and the compatibility in CMOS (Complementary Metal Oxide Semiconductor) technology, its unsatisfying performance still cannot earn it the favor of the mainstream right now. LiDAR upgrading in terms of hardware properties is not as promising as expected in the short run, but that doesn't put LiDAR at its wit's end.

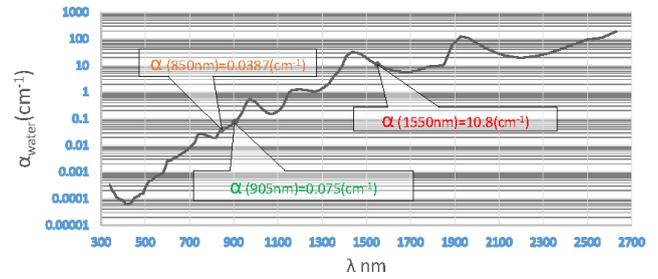

**Figure 22:** Water extinction coefficient spectrum. Laser energy absorption by water of 1550 nm is over 100 times larger than of 905 nm. [364] [365]

### 11.3.2. Dataset and tool support

Based on the weather support status we collected in Table 6, it's easy to get that rain conditions can be considered adequate in current autonomous driving datasets, while fog and snow not so much. Fog or haze is not time-sustained weather that is easy to encounter during data collection, so normally fog datasets are acquired from test facilities or simulators such as those shown in Table 7 and Table 8. As for snow, due to the difference between falling snow and accumulated snow, the qualities of the snow conditions contained in current datasets vary largely, same for simulators. And since the obvious difficulty of constructing artificial snow environment compared to rain, experimental facilities' snow condition supports are very much limited, sometimes are not even exactly designed for AV testings. Furthermore, the strong light and contamination supports are seriously lacking in datasets, even rarer in simulators and facilities, which makes the research on this area slow. Therefore, as rich as the dataset resources are getting, the limitations on weather support are still realistic problems for autonomous driving research in adverse weather.

### 11.4. Future research
### 11.4.1. Other LiDAR types

There is an uncommon LiDAR called super-continuum laser [369], a broadband beam pumped in very short pulse duration. Such technology is widely used in gas sensing, optical communication, etc., and Outsight AI is the company known for developing it in the ADS field [370]. This kind of LiDAR works in the SWIR (Short-wave Infrared) band and can do multispectral detection in real-time. Each wavelength in the SWIR band has a unique reflectance spectral signature based on the object's material, e.g. snow, ice, skin, cotton, plastic, asphalt and so on. That way, it's possible to recognize a real person from a mannequin or a poster, or to classify the ongoing weather.

One hidden problem is that the super-continuum laser promoter, Outsight, does not explain the eye safety conundrum when the wavelength is not 905 nm nor 1550 nm. Considering super-continuum lasers normally work at 728-810 nm [371], the specific information on whether the power level they are using at this wavelength range has any risk to eyes and its performance in field tests with weather presence is not exactly clear. Just like most of the FMCW LiDAR, very few solid-state LiDARs with the ability to compete with mature traditional rotating LiDARs are commercially ready on the market for AVs right now, but mostly for robotics, thus their performances in adverse weather conditions are seldom tested either. That being said, new technologies in new kinds of LiDARs, for example, Baraja's spectrum-scan technology [372] [373], are still being looked forward to.

### 11.4.2. Other camera types

High Dynamic Range (HDR) camera is a type of camera that captures three images of the same scene with three different shutter speeds corresponding to different brightness: bright, medium, and dark. An HDR image that reveals both what's in the dark and glare is then produced by the combination of said three [374]. Clearly, such a feature gives HDR camera a strong advantage in the conditions of strong light or shadows, but it has a serious limitation on moving objects because any movement between successive images will cause a staggered-blur strobe effect after combining them together. What's more, due to the need for several images to achieve desired luminance range, extra time is expected, which is a luxury for video conditions. In order to increase the dynamic range of a video, either the frame rate or the resolution is going to be cut in half for the acquirement of two differently exposed images. If no frame rate or resolution wants to be sacrificed, a CMOS image sensor with dual gain architecture is required. To be of use for ADS, HDR cameras might need some extra calculation algorithms built into the image processing structure. Anyway, each kind of camera has had its own share in ADS during the never-ending attempts of tackling adverse weather.

Hyperspectral imaging technology on the other hand could be the key to the next generation of vision in ADS. Covering an extremely broad spectrum all the way from UV to IR, hyperspectral cameras can record over 100 different wavelengths and filter out visible light interference. The AV situation awareness can be enhanced because this kind of camera can precisely identify the subjects' material signatures by visualizing infrared spectrum. In other words, the material of the detected target can be clearly identified and classified based on its chemical compositions [375]. This distinguish strength makes hyperspectral cameras pretty much immune to all the light related conditions, including darkness, shadow, and direct strong light. Even fog doesn't seem like a serious problem here. The only obstacle between hyperspectral imaging technology and being widely deployed in autonomous driving could be the cost of as high as $20,000-$100,000 USD, while the amount of devices to cover a 360° FOV is not on the low side in the meantime. Good thing is that the reduction of costs is on the horizon and the industry also shows great interest in this promising technology [375]. Basterretxea et al. [308] actually already have presented the HSI-Drive dataset collected by only one 25-band hyperspectral camera containing light condition changes, rainy/wet and foggy conditions across four seasons, as shown in Table 6.

### 11.4.3. Aerial view

Long before being introduced into autonomous driving, LiDAR technology was widely used in geographical mapping and meteorological monitoring. Terrain, hydrology, forestry, and vegetation cover can be measured by LiDAR mounted on planes, or even satellites [376]. The advantage of looking from above is that the view coverage is enormous and there are fewer obstacles than from the ground. With the rapid development of Unmanned Aerial Vehicles (UAV) like drones, it's becoming realistic to do transportation perception from the top view, which sees what couldn't be seen from the ground. For example at an intersection, an AV can only behave according to its leading vehicle but not something that's beyond direct sight, however, UAV can see the leading vehicle of the leading vehicle and thereafter, and foresee risks far away from the subject AV and avoid accidents in advance. Aerial LiDAR makes the AV ego-perception into a macro perspective. Xu et al. [284] developed a method to detect road curbs off-line using aerial images via imitation learning. They take images from the New York City planimetric dataset [377] as input and generate a graph with vertices and edges representing road curbs. There are also mature LiDAR obstacle warning and avoidance systems (LOWAS) for unmanned aerial vehicle sense-and-avoid [285]. LOWAS system is constructed by three key modules: trajectory prediction, potential collision calculation, and optimal avoidance trajectories generation, and they provide the possibility and prove it actionable to use aerial LiDAR in ground transportation navigating.

As far as the authors are concerned, the biggest problem hindering aerial LiDAR from being deployed into ADS right now is transmitting and communicating. Without advanced wireless channels, it's hard to transmit the data or decisions from UAVs to AV in real-time from a certain distance, let alone aerial view is not enough to cover the modern urban environment where tunnels and elevated roads are common. As a result, although researches are being done towards aerial image segmentation and object detection, the main use for aerial LiDAR currently is focused on stationary tasks like structural inspection [286].

## 12. Conclusion

In this work, we surveyed the influence of adverse weather conditions on major ADS sensors like LiDAR and cameras and their influences on the different elements of an AV. Solutions including sensor hardware and mechanical devices were listed. The core solution to adverse driving conditions is perception enhancement and various machine learning and image processing methods like de-noising were thoroughly analyzed. Classification, assessment, control & planning, potential and future auxiliary solutions against adverse weather such as roadside units and V2X were also discussed. A research tendency towards robust sensor fusions, sophisticated networks and computer vision models, and higher support from legislation and public acceptance are concluded. Candidates for future ADS sensors such as FMCW LiDAR, HDR camera and hyperspectral camera were introduced. Finally the limitations brought by the lack of relevant datasets and the difficulty on 1550 nm LiDAR were thoroughly explained. This survey covered almost all types of common weather that pose negative effects to transportation including rain, snow, fog, haze, strong light, and contamination, and listed out datasets, simulators and experimental facilities that have weather support.

With the development of advanced test instruments and new technologies in LiDAR architectures, signs of progress have been largely made in the performance of automated driving in common wet weather. Rain and fog conditions seem to be getting better with the advanced development in computer vision in recent years, but still have some space for improvement on LiDAR. Snow, on the other hand, is still at the stage of dataset expanding and perception enhancement against snow has some more to dig in. Hence, point cloud processing under extreme snowy conditions, preferably with interaction scenarios either under controlled environments or on open roads is part of our future work. Two major sources of influence, strong light and contamination, are still not rich in research and solutions. Anyway, various sensors with their particular strengths in fighting against adverse weather conditions such as thermal cameras and radar are being actively deployed in all kinds of sensor fusion modalities, which send strong reinforcements to the task. Hopefully, efforts made towards the robustness and reliability of ADS can carry autonomous driving research to the next level of autonomy.

## Declaration of Interests

The authors declare that they have no known competing financial interests or personal relationships that could have appeared to influence the work reported in this paper.

## Acknowledgment

Funding: The author (Y. Zhang) would like to take this opportunity to thank the "Nagoya University Interdisciplinary Frontier Fellowship" supported by JST and Nagoya University, and the Core Research for Evolutional Science and Technology (CREST) project of the Japan Science and Technology Agency (JST).

The authors thank Prof. Ming Ding from Nagoya University for his help. We would also like to extend our gratitude to Sensible4, University of Michigan, Tier IV Inc., Ouster Inc., Perception Engine Inc., and Mr. Kang Yang for their support. In addition, our deepest thanks to VTT Technical Research Center of Finland, the University of Waterloo, Pan Asia Technical Automotive Center Co., Ltd, and Civil Engineering Research Institute for Cold Region of Japan.

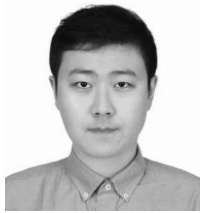

YUXIAO ZHANG received the B.S. degree in mechanical engineering from Wuhan University of Technology, China, and the M.S.Eng from the University of Michigan, USA. From 2019 to 2020, he worked as a Research Assistant at the Integrated Nano Fabrication and Electronics Laboratory of University of Michigan, and as a Graduate Student Instructor with the College of Engineering and Computer Science at the same school. He is currently pursuing the Ph.D. degree with Nagoya University, Japan. His main research interests are LiDAR sensors, and robust perception for autonomous driving systems.

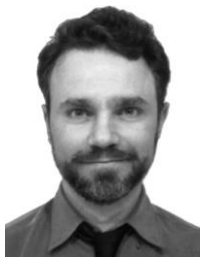

ALEXANDER CARBALLO received the Dr. Eng. degree from the Intelligent Robot Laboratory, University of Tsukuba, Japan. From 1996 to 2006, he was a Lecturer with the School of Computer Engineering, Costa Rica Institute of Technology. From 2011 to 2017, he worked in LiDAR research and development at Hokuyo Automatic Company Ltd. Since 2017, he joined Nagoya University, Japan, where he is currently a Designated Associate Professor affiliated to the Institutes of Innovation for Future Society. He is a professional member of IEEE Intelligent Transportation Systems Society (ITSS), IEEE Robotics and Automation Society (RAS), Robotics Society of Japan (RSJ), Asia Pacific Signal and Information Processing Association (APSIPA), the Society of Automotive Engineers of Japan (JSAE), and the Japan Society of Photogrammetry and Remote Sensing (JSPRS). His main research interests include LiDAR sensors, robotic perception, and autonomous driving.

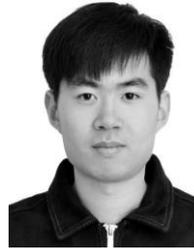

HANTING YANG received his B.S. degree and M.E. degree from Beijing University of Civil Engineering and Architecture. He is currently pursuing a Ph.D. degree with the Graduate School of Informatics, Nagoya University, Japan. His main research interests are image processing, deep learning, and robust vision perception for autonomous vehicles.

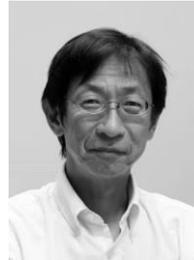

KAZUYA TAKEDA received the B.E. and M.E. degrees in electrical engineering and the D.Eng. degree from Nagoya University, Nagoya, Japan, in 1983, 1985, and 1994, respectively. From 1986 to 1989, he was with the Advanced Telecommunication Research (ATR) Laboratories, Osaka, Japan. He was a Visiting Scientist with the Massachusetts Institute of Technology (MIT), from November 1987 to April 1988. From 1989 to 1995, he was a Researcher and Research Supervisor with the KDD R&D Laboratories, Kamifukuoka, Japan. From 1995 to 2003, he was an Associate Professor with the Faculty of Engineering, Nagoya University. Since 2003, he has been a Professor with Graduate School of Informatics, Nagoya University and currently is the Head of the Takeda Laboratory, Nagoya University. Currently, he also serves as Vice President of Nagoya University. He is a fellow of IEICE (the Institute of Electronics, Information and Communications Engineers) and a senior member of IEEE. Prof. Takeda has served as one of academic leaders in various signal processing fields. Currently, he is a BoG (Board of Governors) member of IEEE ITS Society, Asia-Pacific Signal and Information Processing Association (APSIPA) and vice president of Acoustical Society Japan. He is a co-founder and director of Tier IV, Inc. His main focus is in the field of signal processing technology research for acoustic, speech and vehicular applications. In particular, understanding human behavior through data centric approaches utilizing signal corpora of real driving behavior.